\documentclass[11pt]{article}
\usepackage[T1]{fontenc}
\usepackage{lmodern}
\usepackage[letterpaper,margin=1in]{geometry}
\usepackage{amsmath,amssymb}
\usepackage{graphicx}
\usepackage{booktabs}
\usepackage{array}
\usepackage{xcolor}
\usepackage{tikz}
\usetikzlibrary{arrows.meta,positioning,fit}
\usepackage[numbers,sort&compress]{natbib}
\usepackage{pdfpages}
\usepackage{microtype}
\usepackage[hidelinks]{hyperref}
\graphicspath{{./}}
\title{ChemE-as-Code: An Open-Source Platform for Agentic AI in Chemical Engineering}
\author{Artur M. Schweidtmann\\[4pt]
  \small Process Intelligence Research Group, Department of Chemical Engineering,\\
  \small Delft University of Technology, Van der Maasweg 9, 2629 HZ Delft, The Netherlands\\
  \small \href{mailto:a.schweidtmann@tudelft.nl}{a.schweidtmann@tudelft.nl},
  ORCID \href{https://orcid.org/0000-0001-8885-6847}{0000-0001-8885-6847}}
\date{Preprint submitted to AIChE Journal, November 2026}

\begin{document}
\maketitle

\begin{abstract}
ChemE-as-Code provides an open-source environment in which AI agents built on large language models (LLMs) can carry out chemical engineering work and engineers can check it. Its calculation tools range from catalysis through process simulation to plant layout, HAZOP and life cycle assessment, and skills tell a general-purpose agent how to use them. Engineering files are linked, machine-readable and versioned in Git; calculations run outside the LLM, each value records its origin, and automated checks report inconsistencies and out-of-date results by file and line. In an illustrative ammonia case, agents developed a written task into a preliminary basic-engineering package (simulation files, P\&IDs, layout, HAZOP, datasheets, relief) for USD~224 of API cost in the design sessions. Planted faults and engineering changes show that the checks find what their rules cover and miss the rest, so they guide human review rather than replace it. No independent engineer has reviewed the design.

\end{abstract}

\section{Introduction}
\label{sec:intro}

Generative artificial intelligence (AI) is expected to reshape chemical
engineering~\cite{Schweidtmann2024GenAI,DecardiNelson2024GenAIPSE}. AI agents, its newest form, can go
beyond answering questions: they can use tools, edit files and iterate on their own
results~\cite{Bran2024ChemCrow,Rupprecht2026MultiAgent}. Engineering work can thus increasingly be
proposed, executed and revised by computational agents. In the past years, software development has
shown what this opens up. AI assistants have sped up well-defined programming
tasks~\cite{Peng2023Copilot}, and the length of tasks that agents complete on their own keeps
growing~\cite{Kwa2025LongTasks}. Software teams can use agents in this way because their work already
lives in versioned text files that automated tests can check. We expect chemical process engineering to
benefit in the same way once its work takes the same form.

That form is harder to reach in chemical engineering than in software. A plant is described by many
coupled documents: process models, specifications, process flow diagrams (PFDs), piping and
instrumentation diagrams (P\&IDs), equipment data, layouts, safety studies and environmental
inventories. P\&IDs are the central reference documents of a plant over its life
cycle~\cite{Toghraei2019PIDDevelopment,Towler2008ChemicalEngineeringDesign,ISO10628} and are still
created largely by hand~\cite{Toghraei2019PIDDevelopment,Uzuner2012KnowledgeBasedPID}. The documents
depend on each other, but the dependencies are rarely recorded. This means that a change in one place,
such as a higher design capacity, silently invalidates numbers elsewhere: in the simulation, the
drawings, the equipment data, the layout, the safety analysis and the
inventory~\cite{Clarkson2004ChangePropagation}. Nothing tells the engineer which of them. Much process
information also still sits in documents that must be digitized before a change can be
assessed~\cite{Theisen2023Digitization}. Integrated information models, exchange standards and digital
threads address parts of this
problem~\cite{BayerMarquardt2004DataDocuments,Morbach2009OntoCAPE,Cameron2024DEXPI,Hedberg2016DigitalThread}.
They let tools exchange data. They do not record, for a whole project and at every revision, which
results are still consistent with each other, which are out of date, and who has reviewed what.

AI agents make this problem larger. They produce more proposals and revisions than people can read, and
their output can look plausible but still be invalid. Earlier generators of flowsheets and P\&IDs, from
rule-based systems~\cite{Blitz1994PIDConfiguration,Uzuner2012KnowledgeBasedPID,Obst2013ModularEngineering}
to learned models~\cite{Vogel2023Flowsheets,Hirtreiter2024ControlStructures,SchulzeBalhorn2024FlowsheetAutocorrection},
have a related limit: they produce part of a design without the decisions behind it and outside the
engineering workflow. As agents propose faster, verification increasingly sets the pace of
engineering~\cite{Schweidtmann2026PlantAsCode}.

A further question is how agents should be organized. Our earlier review proposed teams of specialized
agents, one per discipline, mirroring an engineering organization~\cite{Rupprecht2026MultiAgent}. Here we
take a different route. Domain knowledge sits in skills and tools, and one general-purpose agent loads
what a task needs. A skill is a short instruction file that tells the agent how to perform an
engineering task; a tool is a program that performs a calculation or a check. This keeps the engineering
knowledge in files that engineers can read, test and version, independently of a particular agent or
model, and it reduces the need for coordination between discipline-specific agents, a reported source of
failure in multi-agent systems~\cite{Cemri2025MASFail}.
Separate sessions are still used where independence helps, as in review. Which option works better is
an open question that needs a controlled comparison.

\begin{figure}[t]
  \centering
  \includegraphics[width=\textwidth]{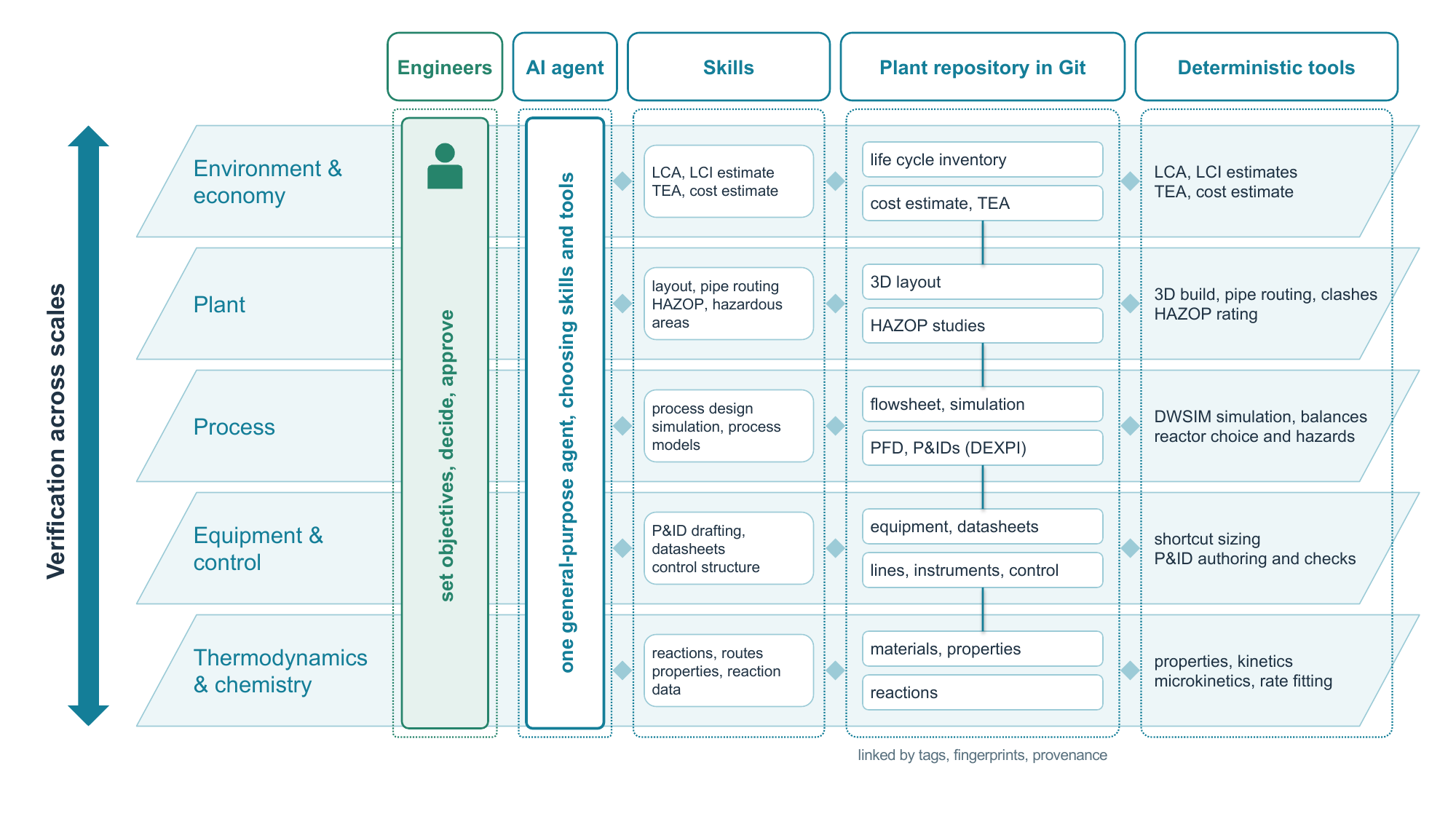}
  \caption{ChemE-as-Code across engineering scales. One general-purpose AI agent selects, by task, the
  skills (instruction files) and the calculation tools of a scale, from chemistry and thermodynamics
  through process and equipment design to plant layout, safety, cost and environmental assessment.
  Engineers (green) set objectives, decide and, in the proposed workflow, approve revisions. All work
  goes into one shared, versioned plant repository in Git. Its files refer to each other by equipment and
  line tags, each calculated value records its inputs, and each value records where it came from, so that
  a change at one scale can be traced to the files it affects at other scales. The domain knowledge lives
  in the skills and tools, not in several specialized agents. Concept adapted from Rupprecht et
  al.~\cite{Rupprecht2026MultiAgent}.}
  \label{fig:multiscale}
\end{figure}

The acceleration that AI agents promise depends on the engineering environment in which they work. We
present ChemE-as-Code, an open-source platform that provides such an environment. It implements the
Plant-as-Code perspective of Schweidtmann and Demeyer~\cite{Schweidtmann2026PlantAsCode} and extends it
from process and plant design to process chemistry, simulation, techno-economic assessment (TEA) and
life cycle assessment (LCA); the name stands for treating chemical engineering work as code. Early
industrial pilots on individual projects have pointed to substantially faster design
iterations~\cite{Schweidtmann2026PlantAsCode}, and ChemE-as-Code makes the approach open, broad and
testable. The platform rests on two foundations. First, a plant is one Git repository of linked,
machine-readable files, from the design basis (the requirements the plant must meet: product, capacity,
feed, site and utilities) to the life cycle inventory. Every revision is checked by programs, and the
checks and calculations behind each value are kept with it; we call this record the evidence of a
revision. Second, an open set of calculation tools and agent skills spans the scales of chemical
engineering, from catalysis and reaction kinetics to plant layout and LCA (Figure~\ref{fig:multiscale}). Agents and
engineers work through the same files and command line, and in the proposed workflow a person reviews
the evidence before a revision is accepted. We argue that the innovation lies in this combination of
capable agents with an engineering environment designed for them. Models will change quickly, but we
expect the way agents interact with engineering files, tools, evidence and reviewers to remain relevant.

We ask three questions. RQ1 (representation): can a process project be represented as linked,
machine-readable files across disciplines that agents and people edit? RQ2 (verification): do automated
checks find inconsistencies, out-of-date results and affected files when the design changes, where do
they fall short, and can generated files be regenerated reproducibly? RQ3 (agent-enabled engineering):
can a general-purpose agent carry out interconnected engineering tasks with these skills and tools, at
what cost and with which limitations? The paper contributes a paradigm for AI-accelerated,
human-directed Chemical Engineering-as-Code, described as five shifts in how engineering work is
organized (Section~\ref{sec:concept}); the open-source platform that implements it
(Section~\ref{sec:platform}); and an end-to-end demonstration on a natural-gas ammonia plant with an
evaluation of what the checks find and miss (Section~\ref{sec:cases}). The case shows that the workflow
runs. It does not establish the quality of the design or a productivity gain, which remains our
hypothesis.

\section{Related Work}
\label{sec:related}
ChemE-as-Code builds on decades of process systems engineering (PSE), engineering information
management and software practice. What it adds is the combination: one versioned project in which agents
make changes and programs check them for people. This section places that combination against four lines
of prior work.

\subsection{Process Engineering Information and Its Management}
Simulation, flowsheeting, optimization and synthesis are the foundations of
PSE~\cite{GrossmannWesterberg2000}, and connecting the information they produce is an old goal of the
field~\cite{MarquardtNagl2004IMPROVE}. Integrated models for data and
documents~\cite{BayerMarquardt2004DataDocuments} and the OntoCAPE
ontology~\cite{Morbach2009OntoCAPE,Wiesner2011InformationIntegration} make process engineering knowledge
machine-readable and reusable. Knowledge graphs now link process knowledge, models and the origin of
data in digital twins~\cite{Akroyd2021UniversalDigitalTwin,Zhang2026KnowledgeGraphDigitalTwin,Alimin2026ChatPID}.
Information technology for integrated process and plant design~\cite{Bramsiepe2014InformationTechnologies},
the digitization of legacy drawings~\cite{Theisen2023Digitization} and the digital thread in
manufacturing~\cite{Hedberg2016DigitalThread} pursue the same aim. Model-based systems engineering and
life cycle standards make traceability and configuration management explicit
practice~\cite{Madni2018MBSE,ISO15288}, and design research has studied how a change to one component
propagates through a product's structure~\cite{Clarkson2004ChangePropagation}.

In industrial practice, engineering document management systems (EDMS) control the revision, approval
status and distribution of documents over a plant's life cycle~\cite{IEC82045}. Data-centric engineering
databases keep P\&IDs, equipment and instrument lists in one store, so that the tags and connections on
a drawing agree with the lists as long as both are edited in that
tool~\cite{Bramsiepe2014InformationTechnologies,Wiedau2019ENPRO}. These systems solve part of the version
and dependency problem. However, an EDMS tracks documents rather than their content: it records which
revision of a P\&ID is current, not whether that revision agrees with the instrument list or the
simulation. Consistency in integrated tools ends at the tool boundary, and much of the information
gathered in design does not reach operation in electronic and interoperable form~\cite{Wiedau2019ENPRO}.

\subsection{Exchange Standards}
P\&ID data usually sit in proprietary formats of computer-aided engineering tools. DEXPI, based on
ISO~15926~\cite{ISO15926}, defines a vendor-independent information model and XML exchange format for
P\&IDs and process models~\cite{Cameron2024DEXPI}. Version 2.0.1 covers block flow diagrams (BFDs),
PFDs and P\&IDs~\cite{DEXPI2026Spec} and is the format ChemE-as-Code stores. pyDEXPI makes DEXPI data
available to Python and to machine learning as graphs~\cite{Goldstein2025PyDEXPI}, so that DEXPI can act
as the interface between generative models and industrial tools; graph learning on DEXPI plant
topologies~\cite{Oeing2023GraphLearning} and large language model (LLM) interaction with
P\&IDs~\cite{Alimin2025TalkingPIDs,Alimin2026ChatPID} build on it. CAPE-OPEN makes unit operations and
property packages interoperable between process simulators~\cite{BelaudPons2002}, AutomationML
exchanges engineering data in automation engineering~\cite{IEC62714}, and building engineering has
open, version-controlled common data environments such as Speckle~\cite{Poinet2020Speckle}. However,
standards solve exchange, not revision consistency. A valid DEXPI file says nothing about whether the
equipment list, the simulation or the HAZOP study still agrees with it. This means that a standard can
carry a design between tools, but it cannot tell an engineer what a change has invalidated.

\subsection{AI in Process Engineering}
AI in process engineering is starting to move from generators toward LLM agents, although these are
still first works, and most methods address one document or one task at a time. The field reaches from
expert systems, including knowledge-based P\&ID
configuration~\cite{Blitz1994PIDConfiguration,Uzuner2012KnowledgeBasedPID,Obst2013ModularEngineering},
through machine learning in
PSE~\cite{Venkatasubramanian2019Promise,Schweidtmann2021MLPerspective,Daoutidis2024MLPSE}, to generative
AI and LLMs~\cite{Schweidtmann2024GenAI,DecardiNelson2024GenAIPSE}, which need to be grounded in domain
knowledge~\cite{Venkatasubramanian2025QuoVadis}. For flowsheets and P\&IDs, prior work has produced text
representations~\cite{dAnterroches2006SFILES,Vogel2023SFILES,Mann2024eSFILES}, generative models that
complete flowsheets and propose control
structures~\cite{Vogel2023Flowsheets,Hirtreiter2024ControlStructures,SchulzeBalhorn2025GraphToSFILES,Stops2023FlowsheetRL},
and rule-based and learned methods that detect and correct
errors~\cite{Dzhusupova2022EngineeringErrors,SchulzeBalhorn2024FlowsheetAutocorrection,SchulzeBalhorn2025RuleBasedPID,SchulzeBalhorn2026UtilityPID}.
Open engineering data are
scarce~\cite{SchulzeBalhorn2023DataAugmentation,Schweidtmann2024MiningProcessInformation}, and generative
models produce topologies without the attributes and decisions behind them, are scored against a single
reference design although several designs can be valid, and are not integrated into engineering
workflows. More recently, LLM agents carry out engineering tasks with tools, from process improvement,
optimization and control~\cite{Lee2024ProcessImprovement,Zeng2025ProcessOptimization,Vyas2025IndustrialControl}
to P\&ID generation~\cite{Gowaikar2024PIDAgent,Srinivas2025AutoChemSchematic}, simulation and model
generation~\cite{Pajak2025MultiAgentLLM,Du2025LLMAgentsSimulation,Rupprecht2025Text2Model,Laub2026TextToFlowsheet,Tian2026TextSimulation},
conceptual separation design~\cite{Rupprecht2026MultiAgent} and question answering on
P\&IDs~\cite{Alimin2025TalkingPIDs,Alimin2026ChatPID}. Most use task-specific tools with little human
oversight~\cite{Rupprecht2026MultiAgent}. Some recent work connects several tasks. Laub et al.\ digitize
process descriptions into flowsheets and build converged Aspen Plus simulations, benchmarked against
flowsheets from 50 experts~\cite{Laub2026TextToFlowsheet}. Tian et al.\ go from textual specifications
to executable process simulations with a multi-agent workflow~\cite{Tian2026TextSimulation}, and Zhang et
al.\ combine a knowledge graph with agents to assemble and calibrate process models for digital
twins~\cite{Zhang2026KnowledgeGraphDigitalTwin}. These works stop at flowsheets, simulations and process
models; they do not manage and verify revisions across the further disciplines of a plant project.
Beyond chemical engineering, tool-augmented agents plan and run chemistry
experiments~\cite{Bran2024ChemCrow,Boiko2023Coscientist}, software-engineering agents work through
agent--computer interfaces~\cite{Yang2024SWEAgent} and are evaluated on real repository
issues~\cite{Jimenez2024SWEBench}, and multi-agent frameworks coordinate several LLM agents on one
task~\cite{Wu2023AutoGen,He2025MASSoftware}.

\subsection{Engineering-as-Code and Reproducibility}
Software engineering manages change through version control, continuous integration and
delivery~\cite{Humble2010ContinuousDelivery}, and Infrastructure-as-Code applies the same practice to
computing infrastructure~\cite{Morris2016IaC}. Engineering-as-Code transfers it to engineering
design~\cite{Song2026EngineeringAsCode}. In plant and automation engineering, industrial practice came
first: Davy Demeyer's Design-Ops work applied Engineering-as-Code and agentic practices before this
work~\cite{Bacidore2025DesignOps,Schweidtmann2026PlantAsCode}, and early industrial pilots on individual
projects point to substantially faster design iterations~\cite{Schweidtmann2026PlantAsCode}. These
pilots use closed software and their details are confidential, so their scope and results cannot be
examined or reproduced.
Plant-as-Code, proposed by Schweidtmann and Demeyer~\cite{Schweidtmann2026PlantAsCode}, provides a
perspective on Engineering-as-Code for process plants.

Engineering results can only be trusted if a reader can see where a value came from and can regenerate
it. Computational science has practices for this. Rules for reproducible
research~\cite{Sandve2013Reproducible} and the FAIR principles~\cite{Wilkinson2016FAIR} describe how to
keep data and code findable and reusable, containers fix the software
environment~\cite{Boettiger2015Docker}, and reproducible builds require that the same inputs yield
outputs that are identical to the byte~\cite{Lamb2022ReproducibleBuilds}. W3C PROV-O provides a
vocabulary for recording which activity and which agent produced a result~\cite{Lebo2013PROVO}. In
engineering projects, by contrast, whether a value was given by a supplier, assumed by an engineer or
calculated by a tool is often recorded only in documents, and a result often cannot be regenerated after
a change.

Taken together, prior work provides rigorous models, shared semantics, portable formats, fast generation
and the software practice for managing change. ChemE-as-Code brings them into one open project in which
every revision is checked across disciplines and kept with its evidence, with tools and skills that let
agents work across the scales of chemical engineering.

\section{The Paradigm of AI-Accelerated Chemical Engineering-as-Code}
\label{sec:concept}
\begin{figure}[t]
  \centering
  \resizebox{\textwidth}{!}{
\begingroup\renewcommand{\familydefault}{\sfdefault}\normalfont\hyphenpenalty=10000\exhyphenpenalty=10000
\definecolor{figink}{HTML}{203344}
\definecolor{figmuted}{HTML}{59717F}
\definecolor{figblue}{HTML}{147E98}
\definecolor{figpale}{HTML}{EDF6F8}
\definecolor{figgray}{HTML}{F3F5F5}
\definecolor{figgreen}{HTML}{26846B}
\definecolor{figred}{HTML}{B5523B}
\begin{tikzpicture}[x=1cm,y=1cm, line cap=round, line join=round, font=\sffamily\scriptsize,
  box/.style={draw=figblue!35, fill=white, rounded corners=3pt, align=center, inner sep=4pt, text=figink,
    text width=1.6cm, minimum height=1.05cm},
  doc/.style={draw=figmuted!60, solid, fill=white, align=center, inner sep=2pt, text=figmuted, font=\sffamily\tiny},
  stage/.style={draw=figblue!50, fill=white, rounded corners=3pt, align=center, inner sep=4pt, text=figink,
    text width=2.35cm, minimum height=1.1cm},
  human/.style={stage, draw=figgreen!70, line width=.8pt},
  change/.style={draw=figred!70, line width=.8pt, fill=white, rounded corners=3pt, align=center, inner sep=3pt,
    text=figink, font=\sffamily\scriptsize\bfseries},
  ask/.style={circle, fill=figred, text=white, font=\sffamily\tiny\bfseries, inner sep=.8pt},
  head/.style={font=\sffamily\footnotesize\bfseries, text=figink, anchor=north west},
  hand/.style={figmuted, line width=.8pt, dashed, -{Latex[length=1.8mm]}},
  flow/.style={figblue, line width=1.1pt, -{Latex[length=2mm]}},
  back/.style={figblue!60, line width=.8pt, dashed, -{Latex[length=1.8mm]}},
  lab/.style={font=\sffamily\scriptsize, text=figmuted, inner sep=1.5pt},
  note/.style={font=\sffamily\scriptsize, text=figmuted, align=left, anchor=north west}]
\fill[figgray,rounded corners=6pt] (0,0) rectangle (7.6,7.0);
\fill[figpale!60,rounded corners=6pt] (7.9,0) rectangle (17.4,7.0);
\node[head] at (0.15,6.85) {(a) Fragmented workflow};
\node[head] at (8.05,6.85) {(b) AI-accelerated ChemE-as-Code};
\node[change, text width=2.4cm] (cha) at (1.3,5.95) {capacity +20\%};
\node[box] (proc) at (1.3,4.45) {\textbf{Process}\\ simulator file};
\node[box] (pid)  at (3.8,4.45) {\textbf{P\&ID}\\ CAE drawing};
\node[box] (eq)   at (6.3,4.45) {\textbf{Equipment}\\ datasheets};
\node[box] (lca)  at (1.3,2.35) {\textbf{LCA}\\ inventory sheet};
\node[box] (haz)  at (3.8,2.35) {\textbf{Safety}\\ HAZOP report};
\node[box] (est)  at (6.3,2.35) {\textbf{Cost}\\ estimate sheet};
\draw[flow, figred!80] (cha) -- (proc);
\draw[hand] (proc) -- node[doc, above=1pt]{PDF} (pid);
\draw[hand] (pid) -- node[doc, above=1pt]{PDF} (eq);
\draw[hand] (eq) -- node[doc, right=1pt]{XLS} (est);
\draw[hand] (pid) -- node[doc, right=1pt]{PDF} (haz);
\draw[hand] (proc) -- node[doc, left=1pt]{XLS} (lca);
\foreach \n in {pid,eq,lca,haz,est} \node[ask] at (\n.north east) {?};
\node[font=\sffamily\scriptsize\bfseries, text=figred, anchor=west, align=left] at (2.75,5.95)
  {which results are still valid?};
\node[note, text width=7.0cm] at (0.3,1.4) {
  $\bullet$ each tool holds its own copy of the data\\
  $\bullet$ dependencies kept in heads and spreadsheets\\
  $\bullet$ out-of-date results are found by hand, if at all};
\node[human] (intent) at (9.35,5.3) {\textbf{Engineer}\\ change: capacity +20\%};
\node[stage] (agent)  at (12.65,5.3) {\textbf{AI agent + tools}\\ edit files, run calculations};
\node[stage] (prop)   at (15.95,5.3) {\textbf{Proposed revision}\\ on a branch, as a diff};
\node[stage] (checks) at (15.95,3.2) {\textbf{Checks}\\ findings, out-of-date results, affected files};
\node[human] (review) at (12.65,3.2) {\textbf{Human review}\\ evidence, then accept or reject};
\node[stage, draw=figgreen!50, fill=figgreen!8] (acc) at (9.35,3.2) {\textbf{Accepted revision}\\ merged with its history};
\draw[flow] (intent) -- (agent);
\draw[flow] (agent) -- (prop);
\draw[flow] (prop) -- (checks);
\draw[flow] (checks) -- (review);
\draw[flow, figgreen] (review) -- node[lab, above]{accept} (acc);
\draw[back] ([xshift=-3mm]checks.north west) -- node[lab, sloped, above, pos=.5]{fix findings} ([xshift=6mm]agent.south east);
\draw[back] ([xshift=-4mm]review.north) -- node[lab, left]{reject} ([xshift=-4mm]agent.south);
\node[draw=figblue!70, line width=.9pt, fill=white, rounded corners=4pt, align=center, text=figink,
  minimum width=8.9cm, minimum height=1.05cm] (repo) at (12.65,0.95)
  {\textbf{Plant repository in Git: the shared engineering state}\\ linked, versioned files; every revision and its evidence kept};
\draw[flow, figgreen] (acc) -- node[lab, right]{merge} (acc |- repo.north);
\draw[figblue!35, line width=.6pt, -{Latex[length=1.6mm]}] ([xshift=-5mm]review.south |- repo.north) -- node[lab, right]{read} ([xshift=-5mm]review.south);
\node[lab, anchor=south east, text=figgreen!80!black] at (17.25,0.06) {green: human authority};
\end{tikzpicture}
\endgroup}
  \caption{The same engineering change in two workflows. (a) Fragmented workflow: a capacity change
  enters one tool, and which downstream results are still valid has to be worked out by hand. Many
  industrial tools are integrated, but the consistency of a whole revision across disciplines is
  difficult to check. (b) AI-accelerated ChemE-as-Code: an engineer states the change, an agent with
  deterministic tools proposes a revision on a branch, the checks report findings, out-of-date results
  and affected files, the agent fixes what it can, and a person accepts or rejects the revision. The
  plant repository holds the shared engineering state, and accepted revisions are merged with their
  history and evidence.}
  \label{fig:workflow}
\end{figure}
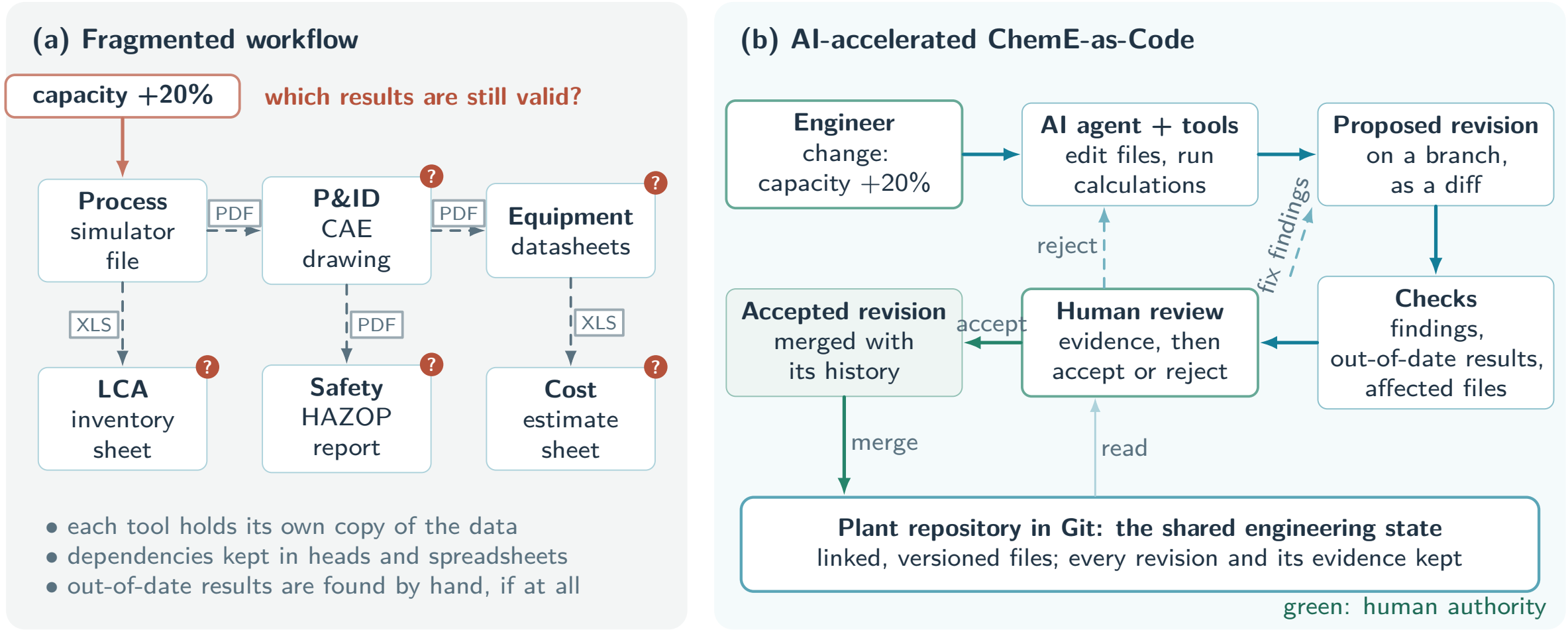

Chemical Engineering-as-Code changes how engineering work is organized, not what engineers calculate.
It extends Plant-as-Code beyond the plant to the whole discipline, including process chemistry,
simulation and LCA. We describe this change as five shifts. Each has a counterpart in software
development~\cite{Humble2010ContinuousDelivery,Yang2024SWEAgent}, which we use as an analogy: what
transfers is the way work is organized, not the nature of the product. Figure~\ref{fig:workflow} shows
one iteration of the resulting workflow for a capacity change, the example we use throughout: the plant
is to make 20\% more product, so the feed, the balances, the equipment, the drawings, the safety study
and the inventory all have to follow.

\medskip\noindent\textbf{(i) From documents to connected engineering knowledge.}
Today, a project is a set of deliverables, each made in its own tool and passed on as a document. In the
new paradigm, the design basis, models, drawings, lists, layouts, safety studies and assessments form
one connected body of machine-readable knowledge, together with the decisions and evidence behind them.
The software analogy is source code and configuration kept as text in one repository, as in
Infrastructure-as-Code~\cite{Morris2016IaC}. The difference is that drawings and documents do not
disappear: they remain as views of the knowledge, issued when a stage requires them. Not every model
has to become executable code, but the representations, decisions and dependencies that engineering
work needs must be readable, changeable and checkable by people and programs alike. For the engineer,
this means that a pump can be followed from the flowsheet to the P\&ID, the equipment
list and the HAZOP study, which an assistant bound to one tool cannot
do~\cite{Schweidtmann2026PlantAsCode}.

\medskip\noindent\textbf{(ii) From manual execution to human-directed agentic engineering.}
Engineers have so far carried out most engineering activities themselves, with tools for individual
calculations. Now they have started to direct some workflows instead. They state objectives,
constraints, specifications and acceptance criteria, and agents draft, calculate, revise and document,
using deterministic engineering tools for every calculated number. The analogy is a developer who
directs coding agents that open pull requests for review. The difference is where responsibility sits:
in engineering, the engineer keeps the responsibility for the design, so an agent's draft is a proposal
until a person accepts it. Engineers can then spend more of their time on alternatives, critical
decisions and review. If a first draft becomes cheap, a broader exploration of design alternatives may
become affordable.

\medskip\noindent\textbf{(iii) From isolated deliverables to continuously evolving projects.}
A deliverable is issued at the end of a phase and revised in the next. In Chemical Engineering-as-Code,
people and agents change a shared project in small revisions. A revision is one recorded change with its
reason, and it can span process, equipment, piping, control, safety and assessment at once. The history
of the design, and why it took its present form, stays with the design. The analogy is small, reviewed
commits with a recorded history instead of a release at the end of a development phase. The difference
is that the engineering phases from concept to detailed engineering remain, with their formal releases
and approvals; what can become much shorter are the iterations within
them~\cite{Schweidtmann2026PlantAsCode}.

\medskip\noindent\textbf{(iv) From periodic checking to continuous verification.}
Checks are traditionally tied to milestones such as design reviews and HAZOP studies. Agents, however,
can propose revisions faster than anyone can check them~\cite{Schweidtmann2026PlantAsCode}, so milestone
checks cannot keep up, and verification has to move into every iteration. A check is a program that
tests one rule over the files, for example that every instrument on a P\&ID is in the instrument list,
and reports a finding with the file and line where the rule fails. Agents repair findings before a
person looks at the revision, and every revision carries its evidence: the results of the checks and the
calculations behind its values. The analogy is automated tests and continuous integration, which run on
every commit. The speed of this loop depends on how cheap, local and trustworthy the checks are. Three
states of work are kept apart. A \emph{proposed} revision is what a person or agent has drafted. A
\emph{checked} revision has no error-level findings; its remaining warnings and the requirements that
cannot yet be checked are stated with it. An \emph{approved} revision has been accepted by a responsible
engineer. The difference from software is firm: checks never replace approval, which is given by people,
and a HAZOP study still needs a team.

\medskip\noindent\textbf{(v) From discipline-specific automation to integrated engineering.}
Instead of automating a flowsheet, a P\&ID or a HAZOP study one at a time, the aim is to connect these
tasks, so that a change and its consequences are followed across disciplines and project stages. The
capacity increase of Figure~\ref{fig:workflow} then reaches the balances, the simulation, the drawings,
the safety studies and the inventory in one revision, and the checks list which files the change
affected, as far as the platform records the links between files. The analogy is one change made across
several services of a software system in one repository and tested as a whole. The difference is that an
engineering change crosses disciplines with their own tools, standards and responsible people, so the
links between files have to be recorded explicitly.

When such projects are open and interoperable, engineering knowledge can accumulate. Designs, models,
decisions and evidence that are today rebuilt within each organization could be inspected, reused and
extended by others, and work at scales that were impractical, such as broad searches over design
alternatives or the reassessment of many designs when a shared assumption changes, could become possible
(Section~\ref{sec:discussion}). These shifts do not depend on a particular technology. Version control,
plain-text formats, exchange standards such as DEXPI and today's agents are implementation choices. The
paradigm is that engineering knowledge is accessible to computational agents in a shared environment
where every action is traceable, every result can be checked, and people decide.

\section{The ChemE-as-Code Platform}
\label{sec:platform}
ChemE-as-Code combines a shared plant repository with calculation tools, skills for agents and automated
checks. Table~\ref{tab:terms} defines the six terms that the rest of the paper relies on.

\begin{figure}[t]
  \centering
  \resizebox{\textwidth}{!}{
\begingroup\renewcommand{\familydefault}{\sfdefault}\normalfont\hyphenpenalty=10000\exhyphenpenalty=10000
\definecolor{figink}{HTML}{203344}
\definecolor{figmuted}{HTML}{59717F}
\definecolor{figblue}{HTML}{147E98}
\definecolor{figpale}{HTML}{EDF6F8}
\definecolor{figgray}{HTML}{F3F5F5}
\definecolor{figgreen}{HTML}{26846B}
\begin{tikzpicture}[x=1cm,y=1cm, line cap=round, line join=round, font=\sffamily\small,
  box/.style={draw=figblue!40, fill=white, rounded corners=3pt, align=center, inner sep=4pt, text=figink,
    minimum height=1.25cm, font=\sffamily\normalsize},
  head/.style={font=\sffamily\normalsize\bfseries, text=figink, anchor=north west},
  flow/.style={figblue, line width=1.2pt, -{Latex[length=2.2mm]}},
  both/.style={figblue, line width=1.2pt, {Latex[length=2.2mm]}-{Latex[length=2.2mm]}},
  lab/.style={font=\sffamily\normalsize, text=figmuted, inner sep=1.5pt}]
\fill[figgray,rounded corners=6pt] (0,7.9) rectangle (17.4,10.25);
\fill[figpale!70,rounded corners=6pt] (0,5.4) rectangle (17.4,7.25);
\fill[figpale!70,rounded corners=6pt] (0,3.0) rectangle (17.4,4.75);
\fill[figgray,rounded corners=6pt] (0,0) rectangle (12.05,2.35);
\draw[figmuted, dashed, line width=.7pt, rounded corners=6pt] (12.6,0) rectangle (17.4,2.35);
\node[head] at (0.15,10.2) {People and agents};
\node[head] at (0.15,7.15) {Shared interfaces};
\node[head] at (0.15,4.65) {Versioned engineering project};
\node[head] at (0.15,2.25) {Deterministic engineering toolkit (\texttt{cheme})};
\node[head, text=figmuted] at (12.75,2.25) {External programs};
\node[box, text width=5.0cm] (agent) at (4.6,8.8) {\textbf{AI agents}\\ load skills, propose and revise changes};
\node[box, draw=figgreen!70, line width=.9pt, text width=4.4cm] (eng) at (12.2,8.8) {\textbf{Engineers}\\ set objectives, review, approve};
\node[box, minimum width=16.6cm, minimum height=0.95cm] (iface) at (8.7,6.05)
  {Git \quad$\cdot$\quad command line \quad$\cdot$\quad continuous integration \quad$\cdot$\quad viewer};
\node[box, draw=figblue!60, line width=.9pt, minimum width=16.6cm, minimum height=0.95cm] (repo) at (8.7,3.65)
  {\textbf{Plant repository in Git:} connected engineering files, their history, decisions and evidence};
\foreach \x/\t in {1.65/{calculations}, 4.55/{consistency checks}, 7.45/{provenance and impact}, 10.35/{gates and maturity}}
  \node[box, text width=2.3cm, minimum height=1.15cm] at (\x,1.0) {\t};
\node[align=center, text=figink, font=\sffamily\normalsize] at (15.0,1.0) {simulation, P\&IDs,\\ properties, LCA};
\draw[flow, figgreen] ([yshift=.22cm]eng.west) -- node[lab, above]{prompt} ([yshift=.22cm]eng.west -| agent.east);
\draw[flow] ([yshift=-.22cm]agent.east) -- node[lab, below]{report} ([yshift=-.22cm]agent.east -| eng.west);
\draw[both] (agent.south) -- node[lab, right]{use} (agent.south |- iface.north);
\draw[both, figgreen] (eng.south) -- node[lab, right]{use, approve, merge} (eng.south |- iface.north);
\draw[both] (8.7,5.575) -- node[lab, right]{edit, commit, read} (8.7,4.125);
\draw[both] (6.0,2.35) -- node[lab, right]{read, write derived files} (6.0,3.0);
\draw[both] (12.05,1.0) -- (12.6,1.0);
\draw[both, rounded corners=4pt] (0.3,1.0) -- (-0.3,1.0) |- node[lab, pos=.25, left, align=right]{commands,\\ findings\\ (file, line)} (iface.west);
\end{tikzpicture}
\endgroup}
  \caption{Architecture of ChemE-as-Code in four layers. Engineers prompt AI agents, which load skills
  and report back, and both work on a versioned plant repository in Git through the same shared
  interfaces (Git, command line, continuous integration, viewer). Engineers accept and approve changes
  (green: human authority, a policy that the sign-off rules support and that branch protection can
  enforce). The deterministic toolkit calculates and checks; it prints every finding with file and line
  on the command line and in continuous integration, where the agent and the engineer read it. For
  simulation, P\&IDs, properties and LCA it calls established external programs (DWSIM, pyDEXPI, thermo
  and CoolProp, Brightway). It contains no LLM calls. SI Section~S2 gives the detailed architecture.}
  \label{fig:architecture}
\end{figure}
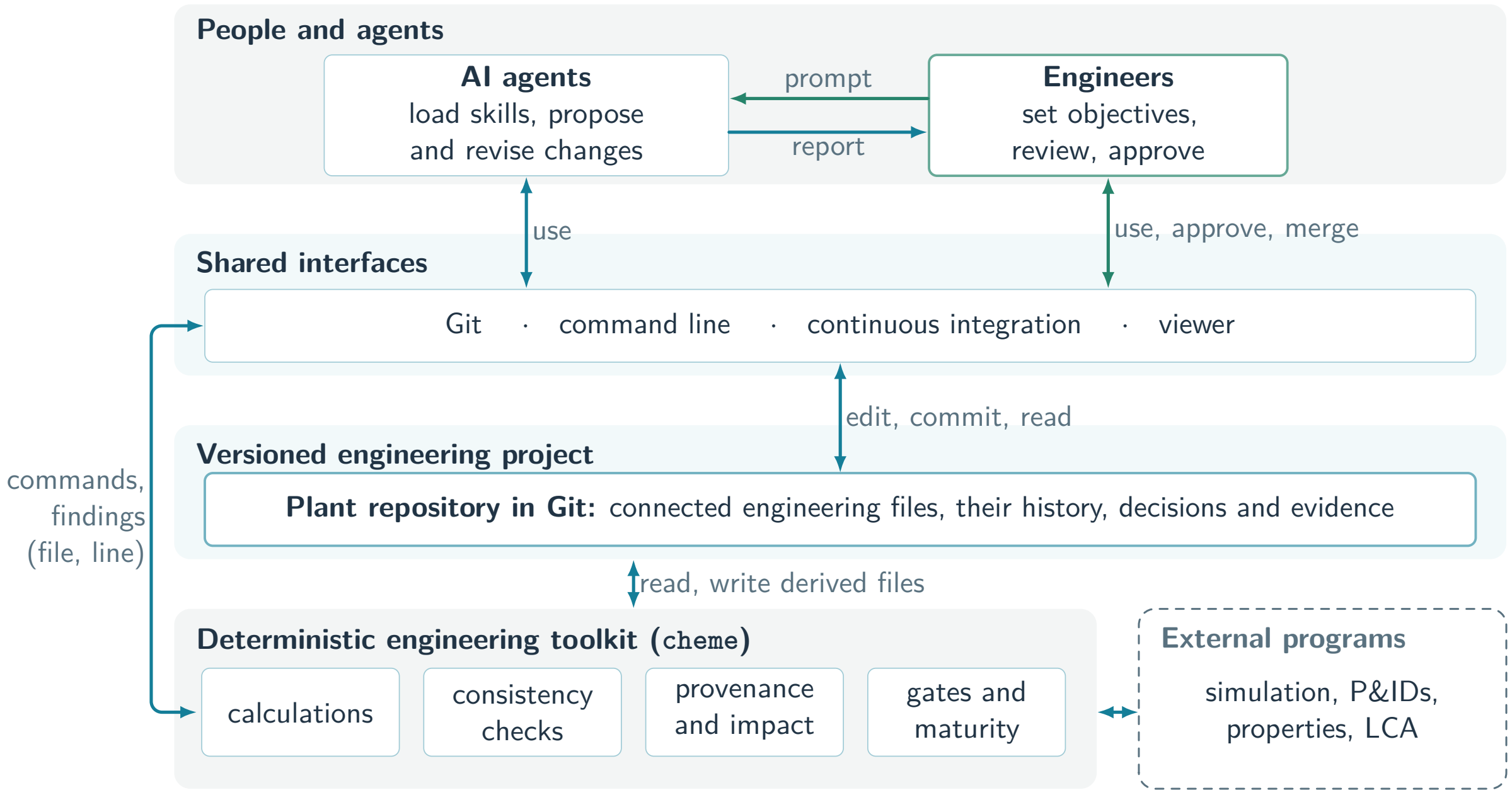

The platform consists of three open repositories. The \emph{core} is the toolkit: its \texttt{cheme}
command validates, formats, renders, compares and generates engineering files, and it includes the
skills that instruct agents and a browser \emph{viewer} (\texttt{cheme serve}). The \emph{project
template} is the starting layout for a new plant repository. The \emph{process library} holds complete
plants, including the case study of this paper. A plant is one Git repository of plain files for every
discipline, from the design basis through process, equipment, piping, control, safety and layout to cost
and the life cycle inventory (LCI), which engineers and agents edit and the core reads, checks and
extends (Figure~\ref{fig:architecture}). This means that an agent sees the whole plant at once, as a
coding agent sees a whole software repository rather than one
file~\cite{Yang2024SWEAgent,Jimenez2024SWEBench}. A change across disciplines is one revision, reviewed
as one diff and reverted as a whole, and the checks compare files of the same revision with each other.
Large software organizations keep most of their code in a single repository for related reasons, namely
unified versioning and atomic changes across projects~\cite{Potvin2016SingleRepository}. At version~1.1.0
the core comprises about 143,000 lines of Python with about 3,000 test functions. SI Section~S2 gives
its detailed architecture, and SI Section~S4 lists its tools and skills by scale.

\begin{table}[!htbp]
  \centering
  \caption{Six terms used in this paper, with the software practice each one resembles. The analogy
  transfers the way work is organized, not the engineering content. A full glossary is given in SI
  Section~S4.}
  \label{tab:terms}
  \footnotesize\renewcommand{\arraystretch}{1.15}
  \begin{tabular}{>{\raggedright\arraybackslash}p{0.17\columnwidth}>{\raggedright\arraybackslash}p{0.55\columnwidth}>{\raggedright\arraybackslash}p{0.21\columnwidth}}
    \toprule
    \textbf{Term} & \textbf{Meaning in this paper} & \textbf{Software analogy} \\
    \midrule
    Check, finding & A program tests one rule over the files; where the rule fails, it reports a finding with the file, the line, a severity and a hint & Automated test; failing test \\
    Skill & A short instruction file that tells an agent when and how to perform an engineering task, which tools to use, and what is left to people & Runbook, contributor guide \\
    Tool & A deterministic program that calculates or checks; it never calls an LLM & Library, compiler \\
    Provenance & The record of where a value came from: given (with a source), assumed (with a reason) or calculated (by a named tool) & Lock file that records the origin of each dependency \\
    Fingerprint; out of date & A hash of a calculated value's inputs; when the inputs change, the value is out of date until it is recalculated & Build cache, invalidated when a source changes \\
    Gate, sign-off & A gate is a set of checks a project stage must pass; a sign-off is the recorded acceptance of a stage by a reviewer & Required status check; merge approval \\
    \bottomrule
  \end{tabular}
\end{table}

\subsection{Engineering Files in Git}
Git keeps versions of files and records who changed what, when and why. It does not know what the files
mean. Git does not distinguish a pump from a pipe, and it merges text without regard to engineering
meaning, so two revisions that are each consistent can merge into one that is not. Suppose one revision
renames pump P-301A to P-303A in the equipment list, and another, made in parallel, adds a line to
P-301A on the P\&ID. Git merges both without complaint, and the merged state names a pump on the P\&ID
that the list no longer has. ChemE-as-Code adds the engineering meaning that Git lacks. Each file
declares its kind and the schema it follows, items refer to each other by tag (the number of a piece of
equipment or a line, such as H-101 or NG-1003), and the checks run on every proposed revision in
continuous integration. In the example, \texttt{cheme validate} reports that the P\&ID refers to a tag
that is not in the equipment list, with the file and the line, before the revision can be accepted.

The following standards and formats are used. P\&IDs are stored in DEXPI~2.0.1~\cite{Cameron2024DEXPI},
based on ISO~15926, and handled with pyDEXPI~\cite{Goldstein2025PyDEXPI}; the BFD and PFD are generated
as DEXPI process models from the flowsheet. All other engineering information is held in YAML files,
each validated against a versioned JSON Schema that the file declares; the declaration is added to older
files by \texttt{cheme migrate}, the mechanism through which files will move to later schema versions.
Decisions are kept in a decision log (\texttt{DESIGN.md}), reviews as Markdown records, and stage
sign-offs in a generated status file. Three properties make these files suitable for agents and for
review. They are \emph{canonical}: \texttt{cheme fmt} writes DEXPI XML with a readable identifier on
every object, so that a P\&ID sheet of tens of thousands of lines still yields a small, reviewable diff.
They are \emph{linked}: list items refer to P\&ID objects by their tags, flowsheet nodes link to
equipment and sheets, and HAZOP studies record the revision of the sheets they analyzed. Relations are
recorded only where a check needs them, not in a universal object model, so an ontology, a knowledge
graph or an EDMS can be built from a plant repository. They are \emph{typed}: every YAML kind has a
schema, so a wrong unit or a missing field is a finding, not a surprise downstream.

The files at one commit are the engineering state of the plant and its source of truth. A person or an agent changes it by
proposing a revision, normally on a branch, which becomes the next state once accepted in review and
merged. In the proposed workflow, acceptance is a human decision, and the change is merged by the
reviewer, not by its author; the platform does not enforce this yet, but branch protection on the Git
host can. In the case study (Section~\ref{sec:cases}), a reviewer agent accepted
the changes, so the design is agent-generated, with the author's recorded interventions and no
independent human review. Such an acceptance lets a change merge; it is not the approval of a
responsible engineer.

Every file is either \emph{authoritative} or \emph{derived}. Authoritative files are written by people
or agents: the design basis, given and assumed values, P\&IDs, lists, layout inputs, HAZOP records,
decisions and reviews. Derived files are generated by a tool: the BFD and PFD from the flowsheet,
simulation runs and calculated values, the 3D plant model, the material take-off (MTO), estimates and
LCI datasets. Derived files are never edited by hand. Where implemented, they record a fingerprint of
their inputs and the version of the tool that generated them, and \texttt{cheme validate} reports a
file or value whose inputs no longer match as out of date, as it reports a hand edit of a generated
sheet. A calculated value never overwrites a given or assumed one: a tool writes only the values it
calculated itself, so if a given flow no longer closes a balance, a check reports the imbalance instead
of a tool correcting the input. Nothing is regenerated, rewritten or declared invalid automatically. A
derived file stays as it is, marked out of date, until a person or agent runs the command that
regenerates it, and regenerating it with the same inputs and the same tool version gives output that is
identical to the byte (SI Section~S4).

\subsection{Tools and Skills for Agents}
The core separates what agents propose from what tools compute. Its engineering domains (chemistry,
process, equipment, piping, control, safety, spatial and assessment) contain no calls to LLMs.
Calculated values enter the files only through deterministic commands or recorded scripts; in the case
study, the agents wrote some of these scripts themselves, which Section~\ref{sec:discussion} discusses.
Established programs do the heavy calculations: DWSIM solves the flowsheet, thermo, chemicals and
CoolProp supply physical properties, pyDEXPI reads and writes P\&IDs, and Brightway computes the
LCA~\cite{Mutel2017Brightway}. The core's own tools cover what sits between them: balances and
propagation, shortcut sizing of exchangers, pumps, compressors and columns, the 3D model with pipe
routing and clash detection, HAZOP rating, cost estimation from published correlations, and TEA. SI
Section~S4 lists them by scale.

The agent's interface is the command line and the files themselves, the same interface a software
engineer uses. The loop is simple. The agent edits a file, runs \texttt{cheme fmt} and \texttt{cheme
validate}, reads the findings, fixes what they report, and validates again. Each finding is one line
with a stable identifier, the rule, a severity, the file and, where one applies, the line, and a hint.
An agent can therefore act on a finding as it would on a failing test. The engineer sees the same lines
in continuous integration and in the viewer.

Skills carry the engineering procedure. A skill is a short instruction file in the plant repository that
tells the agent which inputs a task needs, how to carry it out, which tools to call, and which decisions
are left to people. The 109 skills of
version~1.1.0 span the scales of a project, from component properties and reaction kinetics through
thermodynamic-model selection, flowsheet simulation and reactor screening to equipment sizing, P\&ID
authoring, control, 3D layout, HAZOP, cost estimation, TEA and LCA. About half of them are process
knowledge: unit operations, utilities, heat integration, and the technologies of gas processing,
refining, petrochemicals and heavy chemicals.

The skill for selecting a thermodynamic model shows the division of labor between skill, tool and
person. It is used before any phase calculation and before a property method is recorded in the
materials file. The skill tells the agent to establish the facts first: the pressure range, whether the
mixture is polar, whether electrolytes are present, whether interaction parameters are available, and
whether a second liquid phase or vapor association is expected. The agent fills in only what the given
information or well-established chemistry supports and leaves the rest unset. The tool behind the skill
is a rule engine that implements Carlson's published decision tree. If a fact it needs is missing, it
refuses and names the missing fact as a question, which the agent answers or escalates. The result is a
recommendation with its reasoning, recorded with provenance, and the skill states that it is not a
validated choice. In the ammonia case, this skill recommended a predictive equation of state or a cubic
equation of state with Huron--Vidal mixing rules for the polar loop mixture at 155~bar. The simulation
route could not run either, so the agents used Peng--Robinson as a substitute and recorded the decision,
and the calculation maturity later reported these mixtures as outside the benchmarked range of the
method (SI Section~S8). The HAZOP skill works in the same way: it drafts nodes, deviations and
safeguards, the rating tool looks up the risk band in the matrix, and every entry stays marked as
proposed until a HAZOP team accepts it.

This division means that engineering expertise becomes accessible to agents without being embedded in
an LLM. The numbers come from tools that can be tested and versioned. The procedures come from skills
that engineers can read, review and improve like any other file. The decisions stay with people.

\subsection{Checks, Gates and Human Approval}
\label{sec:verification}
Agents produce more revisions than an engineer can read, so the checks tell the engineer where to look
and what not to trust yet. Each check tests one rule that a program can decide, in four
layers~\cite{Schweidtmann2026PlantAsCode}: every file matches its schema (a design temperature is a
number with a unit); files agree with each other (a line's diameter in the line list matches the P\&ID
that draws it); balances and simulations hold (every mass balance closes within its tolerance); and
selected safety requirements are met (a trip does not share its transmitter with the control loop it
protects against). The core defines 466 rules (SI Section~S3), and every finding names the file and,
where one applies, the line.

A design change shows how the checks work together. Suppose an agent raises the design capacity by
20\% and the feed flows with it. The balance rules report that the flowsheet no longer closes; every
calculated result whose input dependencies are recorded, from stream properties to the simulation run,
is reported as out of date through its fingerprint; and \texttt{cheme impact} lists the files and
reviews that need attention. Nothing is recalculated on its own. The agent or the engineer reruns the
relevant tools and checks whether the affected results are now current, and the reviews that the
change made out of date are redone. This sequence
(check, identify the affected work, recalculate, review) follows only the dependencies the platform
records; a duty copied by hand from the flowsheet into a datasheet is not among them
(Section~\ref{sec:evaluation}).

The checks support the decision whether work can proceed to the next project stage; they do not take
it. Each stage has gates (Table~\ref{tab:terms}), for example that every item on the equipment list has
a datasheet marked final, and a stage is signed off only when its gates pass and no serious review
finding is open (SI Section~S3). The platform also reports calculation maturity, a program-evaluated
summary of which calculation requirements have been met (SI Section~S3). Passing these requirements
establishes only that the conditions encoded in them have been met, not that the design is right or
that an engineer has approved it. Approval is reserved for people in
the proposed workflow; the core records reviews but does not enforce this, and in the case study it
accepted sign-offs on agent reviews.

\section{Ammonia Case Study}
\label{sec:cases}
We demonstrate the platform on a natural-gas ammonia plant, following the design from its initial
requirements through simulation, engineering documents and safety studies. We chose ammonia because its
design couples many disciplines, from reaction equilibria and recycle convergence through equipment,
piping, instrumentation and layout to cost estimation. We ask
whether agents can use the platform to develop a connected engineering project (RQ1, RQ3), how far the
checks identify errors and incomplete changes, whether derived files can be regenerated (RQ2), and
whether an agent can revise the plant through the same workflow (RQ2, RQ3). Three kinds of success have
to be kept apart: execution, meaning that agents generated and modified a multidisciplinary project;
formal verification, meaning that the implemented rules and gates checked the conditions they cover; and
engineering correctness, which this case does not establish, because no independent engineer has
reviewed the design.

\subsection{Setup}
\label{sec:ammonia}
Agents developed the plant in two stages (Table~\ref{tab:case}): a concept stage of three sessions that produced the design basis, a balanced flowsheet and the equipment list, and an
engineering stage that started from a written brief by the author (SI Section~S5). In the engineering
stage, an orchestrating session ran one implementation job per section, each followed by a reviewer
job in a separate session that returned the work for correction or merged it once continuous
integration had passed. These are workflow roles of one general-purpose agent, not discipline-specific
agents. Unlike in the proposed workflow, reviewer agents accepted the changes, so no engineer's
acceptance enters the record. The author's interventions were the brief, the budget and stop rules,
and the decisions the agents escalated (SI Section~S9). The evaluations of
Section~\ref{sec:evaluation} ran on the reported state, process-library tag \texttt{nh3-run-final},
where every file of the case is public.

\begin{table}[!htbp]
  \centering
  \caption{The ammonia case study at a glance. Sources: the run records and the engineering files at
  process-library tag \texttt{nh3-run-final}; details in SI Sections~S5 to~S10.}
  \label{tab:case}
  \footnotesize\renewcommand{\arraystretch}{1.15}
  \begin{tabular}{>{\raggedright\arraybackslash}p{0.2\columnwidth}>{\raggedright\arraybackslash}p{0.75\columnwidth}}
    \toprule
    Plant & 1,500~t/d of refrigerated liquid ammonia from natural gas in the Rotterdam area: desulfurization, prereforming, fired primary and air-blown secondary reforming, water-gas shift, CO$_2$ removal, methanation, drying, 150~bar synthesis loop with refrigeration and purge-gas recovery; electric drives, steam export \\
    Input from the author & Concept stage: a task and a literature selection in a process-library issue. Engineering stage: a written brief (``take ammonia-1 to a complete design: simulation, basic engineering, HAZOP, datasheets, relief''), budget caps and stop rules \\
    Stages & Concept stage, 27 October 2026: 3 sessions, 40~min. Engineering stage, 29 to 30 October 2026: 60 sessions, about 27~h \\
    Agent and roles & Claude Opus~5.5 in every session; the roles implement, review, fix, sign off and merge differ only by instructions, section task and loaded skills \\
    Human interventions & Eight in the engineering stage, counting the brief: budget and stop rules, two platform bug fixes outside the design, and the decisions the agents escalated \\
    Output & Process: flowsheet and converged DWSIM simulation. Engineering: DEXPI P\&ID sheets, equipment, line and instrument lists, control loops, datasheets, routed 3D layout. Safety: HAZOP study, relief cases. Assessment: AACE class~4 estimate, TEA, LCI. Records: decisions and reviews (counts in SI Section~S9) \\
    Checks at the reported state & 0 errors, 268 warnings; all seven concept and 13 basic gates pass, three of five detail gates; calculation maturity \emph{balanced}, the second of four levels (SI Section~S3) \\
    Cost & USD~255 of recorded API cost for 63 sessions, of which USD~224 in the 56 design sessions (Table~\ref{tab:cost}) \\
    Reviewed by & Reviewer agents of the same model; no independent engineer \\
    \bottomrule
  \end{tabular}
\end{table}

\subsection{The Engineering Package}
\label{sec:project}
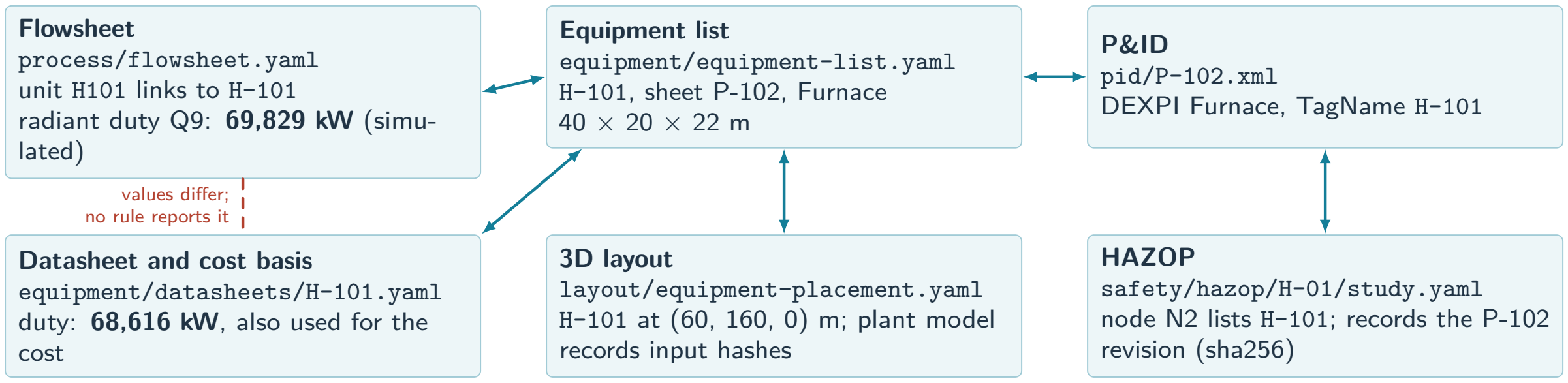
\begin{figure}[!p]
  \centering
  \begin{minipage}[b]{0.58\textwidth}\centering
    \includegraphics[width=\textwidth, trim=0 222 0 90, clip]{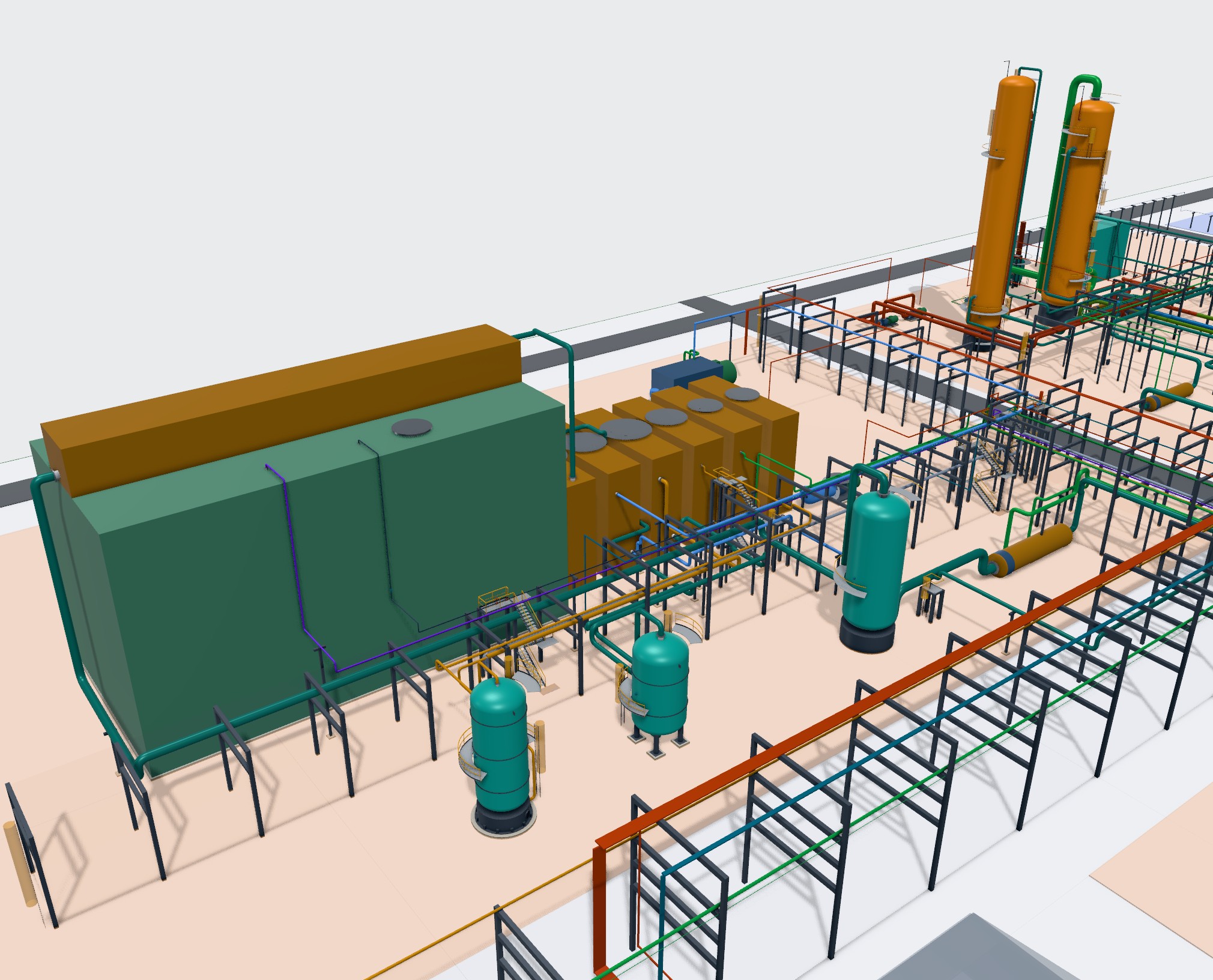}\\[2pt]{\footnotesize (a) 3D plant model: H-101 (green block), its convection section (brown blocks) and the surrounding plant}
  \end{minipage}\hfill
  \begin{minipage}[b]{0.40\textwidth}\centering
    \includegraphics[width=\textwidth, trim=171 85.9 506.55 270, clip]{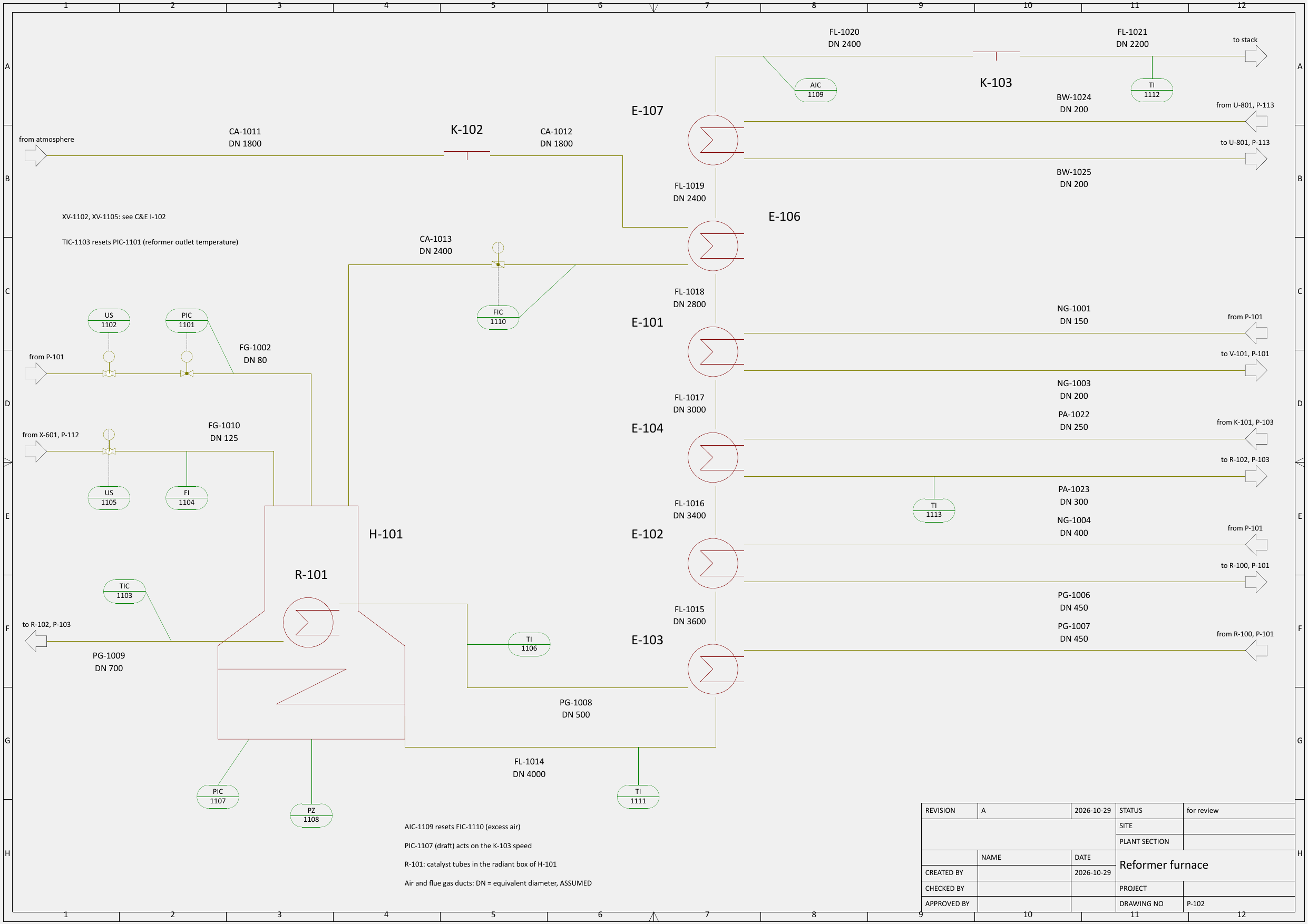}\\[2pt]{\footnotesize (b) P\&ID P-102, excerpt around H-101, R-101 and the exchangers}
  \end{minipage}\\[5pt]
\begingroup\scriptsize
\definecolor{figink}{HTML}{203344}
\definecolor{figblue}{HTML}{147E98}
\newcommand{\figtag}[1]{\textcolor{figblue}{\textbf{#1}}}
\color{figink}\renewcommand{\arraystretch}{1.12}\setlength{\tabcolsep}{3pt}
\begin{tabular}{>{\raggedright\arraybackslash}p{0.10\textwidth}>{\raggedright\arraybackslash}p{0.19\textwidth}>{\raggedright\arraybackslash}p{0.195\textwidth}>{\raggedright\arraybackslash}p{0.205\textwidth}>{\raggedright\arraybackslash}p{0.105\textwidth}>{\raggedright\arraybackslash}p{0.13\textwidth}}
\toprule
\textbf{Deviation} & \textbf{Cause} & \textbf{Consequence} & \textbf{Safeguards} & \textbf{Risk: initial, residual} & \textbf{Actions} \\
\midrule
S5: More temperature (\figtag{R-101} tubes) & TIC-1103 fails low or reads low (or PIC-1101 fails, PV-1101 sticks open); \figtag{H-101} is over-fired & \figtag{R-101} tube walls exceed their design temperature; tube rupture; process gas burns in the firebox & SG7 TIC-1103 high alarm (not credited, part of the causing loop); SG8 TIC-1103 resetting PIC-1101 & people 3/C high, asset 4/C high; residual not rated & A6 burner management; A7 independent trip, tube skin temperature \\
\addlinespace
S7: less combustion air, draft & ID fan K-103 trips or its damper closes; \figtag{PIC-1107} holds the last speed; furnace at positive pressure & Hot flue gas and flames at peepholes and openings; burns to people on the burner platforms; incomplete combustion & SG9 \figtag{PIC-1107} high pressure alarm; SG10 interlock I-102: \figtag{PZ-1108} high-high closes XV-1102 (fuel gas) and XV-1105 (tail gas) & people 3/C high; residual people 3/B medium & none \\
\bottomrule
\end{tabular}
\endgroup
\\[2pt]{\footnotesize (c) HAZOP node N2, two scenarios shortened from the worksheet; deviations are HAZOP guideword phrases; proposed by agents, not reviewed by a team; tags in blue appear in (b)}\\[5pt]
  \resizebox{\textwidth}{!}{
\begingroup\renewcommand{\familydefault}{\sfdefault}\normalfont
\definecolor{figink}{HTML}{203344}
\definecolor{figmuted}{HTML}{59717F}
\definecolor{figblue}{HTML}{147E98}
\definecolor{figpale}{HTML}{EDF6F8}
\definecolor{figred}{HTML}{B3402E}
\begin{tikzpicture}[x=1cm,y=1cm,line cap=round,line join=round,font=\sffamily\scriptsize,
  box/.style={draw=figblue!40,fill=figpale,rounded corners=2pt,align=left,inner sep=4pt,text=figink,text width=4.9cm,minimum height=1.55cm,anchor=north west},
  link/.style={figblue,line width=.9pt,{Latex[length=1.6mm]}-{Latex[length=1.6mm]}},
  rule/.style={font=\sffamily\tiny,text=figmuted,midway,align=center},
  gap/.style={figred,line width=.9pt,dashed}]
\node[box] (fs) at (0,3.4) {\textbf{Flowsheet}\\ \texttt{process/flowsheet.yaml}\\ unit \texttt{H101} links to \texttt{H-101}\\ radiant duty Q9: \textbf{69,829 kW} (simulated)};
\node[box] (eq) at (5.9,3.4) {\textbf{Equipment list}\\ \texttt{equipment/equipment-list.yaml}\\ \texttt{H-101}, sheet P-102, Furnace\\ 40 $\times$ 20 $\times$ 22 m};
\node[box] (pid) at (11.8,3.4) {\textbf{P\&ID}\\ \texttt{pid/P-102.xml}\\ DEXPI Furnace, TagName \texttt{H-101}};
\node[box] (ds) at (0,0.9) {\textbf{Datasheet and cost basis}\\ \texttt{equipment/datasheets/H-101.yaml}\\ duty: \textbf{68,616 kW}, also used for the cost};
\node[box] (lay) at (5.9,0.9) {\textbf{3D layout}\\ \texttt{layout/equipment-placement.yaml}\\ \texttt{H-101} at (60, 160, 0) m; plant model records input hashes};
\node[box] (hz) at (11.8,0.9) {\textbf{HAZOP}\\ \texttt{safety/hazop/H-01/study.yaml}\\ node N2 lists \texttt{H-101}; records the P-102 revision (sha256)};
\draw[link] (fs.east) -- (eq.west);
\draw[link] (eq.east) -- (pid.west);
\draw[link] (eq.south) -- (lay.north);
\draw[link] (pid.south) -- (hz.north);
\draw[link] (eq.south west) ++(0.4,0) -- (ds.north east);
\draw[gap] (fs.south) -- (ds.north) node[midway,left,font=\sffamily\tiny,text=figred,align=right] {values differ;\\no rule reports it};
\end{tikzpicture}
\endgroup}\\[2pt]{\footnotesize (d) H-101 across the linked files; solid arrows are references checked by rules}
  \caption{The reformer furnace H-101, one piece of equipment, followed through the files that describe
  it: the 3D model (a), the DEXPI P\&ID (b), the HAZOP study (c) and the links between the files (d).
  Panel (d) also shows a limit of the checks: rules check that a tag refers to the same item across
  files, while a duty that differs between two files is not reported. All drawings, the full HAZOP
  worksheet source and further 3D views are given in SI Sections~S6 and~S10 or the repository. The
  design has not been reviewed by independent engineers.}
  \label{fig:case}
\end{figure}

The engineering stage turned the concept-stage flowsheet into a connected package: a converged
simulation, P\&IDs with their lists, control loops and datasheets, a routed 3D layout, a HAZOP study,
relief cases, a class~4 estimate with TEA, and an LCI, in one repository whose files refer to each other
by tag (Table~\ref{tab:case}). It passed every concept- and basic-stage gate and three of the five
detail gates with no error-level finding. Figure~\ref{fig:case} follows one piece of equipment, the
reformer furnace H-101, through these files: the 3D model, the P\&ID, the HAZOP study and the links
between them. Rules check that its tag refers to the same item in the flowsheet, the equipment list, the
P\&ID, the datasheet, the layout and the HAZOP study, and the HAZOP study records the revision of the
sheet it analyzed, so a later change to the sheet marks the study as needing review. The figure also
shows a limit of these links. The datasheet and the cost basis state a duty of 68,616~kW, whereas the
flowsheet changed later and now gives 69,829~kW; no rule compares the two, so no finding reports the
difference.

The techno-economic assessment gives a levelized cost of 688~EUR/t against 439~EUR/t assumed for
imported ammonia. Under its assumptions, 2025 Dutch industrial energy prices and a US Gulf Coast import
price without freight (SI Section~S7), the plant would not compete with imports. The LCI records every
exchange of the flowsheet per kg of ammonia and gives a gate-to-gate carbon footprint of 1.46~kg
CO$_2$-eq per kg, with 6\% of the burden allocated to the exported steam by exergy. This counts only
the plant's own emissions: the background datasets for natural gas supply, electricity and utilities
are deliberately left unlinked as open gaps, so the figure is a screening result, not a cradle-to-gate
footprint.

The package is preliminary, not an engineering-approved basic design. The basic gates passed only after
the author accepted two layout exceptions that the agents had escalated rather than decided (SI
Section~S9), the stage sign-offs rest on agent review records, and review findings remain open. The
calculation maturity reaches \emph{balanced}, not
\emph{predictive}: several reactors and separators use specified conversions and splits, the furnace,
refrigeration and steam system are evaluated by agent-written scripts on simulated duties, and the
property method has not been benchmarked for most of the mixtures (SI Sections~S7 and~S8). Three of the
eight relief valves still lack a required capacity, and every HAZOP entry is an agent's proposal,
because no HAZOP team has met.

Reviewer agents recorded 2 high, 39 medium and 52 low findings in 26 review sessions, and 13 times over
7 review rounds they sent work back for fixing. The findings included engineering defects as well as
formal ones: exchangers with a large pressure step between their sides had no tube-rupture case, and
heat-transfer coefficients for gases near atmospheric pressure had been used for the 150~bar loop. As
the brief required, eight jobs stopped and reported a blocker instead of relaxing a gate. One job,
however, marked datasheets final so that a gate would count them, and a reviewer caught this. The run
also found eight defects in the platform software, of which three were fixed by agent jobs outside the
design (SI Section~S9). These reviewers use the same model as the designers and are therefore not
independent~\cite{Schweidtmann2026PlantAsCode}.

\subsection{Cost and Effort}
\label{sec:cost}
The recorded run used 63 agent sessions at an API cost of USD~255.31 (Table~\ref{tab:cost}): USD~12.02 for the concept stage (3 sessions, 40 minutes) and USD~243.30 for the engineering
stage (60 sessions, about 27 hours). Layout and its downstream sections took 40\% of the design cost and
26 of the 56 design sessions, because they needed the most iteration. Review and fix jobs took about a
quarter of the design cost, and nearly half when the close-out jobs (USD~44), which are booked as
implementation, are included. These figures describe one execution with a particular model and heavy
caching (565 million cached input tokens), on a first-generation toolchain and with a workflow nobody
had practiced before. They bound neither the best nor the typical case, and a more mature platform and
more practiced briefs will change them. An order-of-magnitude estimate of the inference energy from the
recorded tokens and published per-token figures~\cite{Oviedo2025EnergyInference,Vartziotis2026TokensWattHours}
is given in SI Section~S9. The cost figures exclude the orchestrating session, whose own API cost was not
recorded, as well as human expertise, infrastructure and independent review. They are not evidence of
productivity compared with an engineering team.

\begin{table}[!htbp]
  \centering
  \caption{Recorded API cost of the ammonia run per section in US dollars, split into implementation jobs and the
  review, fix, sign-off and merge jobs that followed them. Values are rounded. Data: the run ledger at
  \texttt{nh3-run-final}; per-session values in SI Section~S9.}
  \label{tab:cost}
  \footnotesize
  \begin{tabular}{>{\raggedright\arraybackslash}p{0.46\columnwidth}rrrr}
    \toprule
    \textbf{Section} & \textbf{Sessions} & \textbf{Implementation} & \textbf{Review} & \textbf{Total} \\
    \midrule
    Concept stage: design basis, balanced flowsheet, preliminary LCI & 3 & \$8.71 & \$3.30 & \$12.02 \\
    S1 Concept close-out and sign-off & 5 & \$6.50 & \$2.61 & \$9.11 \\
    S2 Process simulation (DWSIM) & 3 & \$28.03 & \$2.41 & \$30.44 \\
    S3 P\&IDs, 14 sheets & 6 & \$28.04 & \$6.59 & \$34.63 \\
    S4 Control and lists & 2 & \$5.07 & \$1.95 & \$7.01 \\
    S5 Layout, MTO, estimate, TEA, basic sign-off & 26 & \$57.26 & \$32.23 & \$89.50 \\
    S6 HAZOP & 4 & \$6.12 & \$5.71 & \$11.83 \\
    S7 Datasheets & 2 & \$6.36 & \$1.73 & \$8.09 \\
    S8 Relief & 2 & \$5.06 & \$1.70 & \$6.76 \\
    S9 Detail gates & 3 & \$12.43 & \$1.68 & \$14.11 \\
    \midrule
    Design, subtotal & 56 & \$163.58 & \$59.92 & \$223.51 \\
    Platform software fixes (outside the design) & 6 & \$20.13 & \$9.55 & \$29.68 \\
    Key performance indicators & 1 & \$2.13 & -- & \$2.13 \\
    \midrule
    Total & 63 & \$185.84 & \$69.47 & \$255.31 \\
    \bottomrule
  \end{tabular}
\end{table}

\subsection{Plausibility Against Open References}
\label{sec:kpis}
To obtain an outside reference, we compared eight of ten planned indicators with the EU reference
document on best available techniques (BAT) for ammonia (BREF)~\cite{EC2007BREFLVIC} and the
International Energy Agency (IEA) ammonia roadmap~\cite{IEA2021AmmoniaRoadmap} (Table~\ref{tab:kpis};
the two NOx indicators are not modeled, and the net energy appears twice, as delivered and with
electricity at an assumed 40\% conversion). The reference values were written before the engineering stage and withheld
from the design jobs. Overall, the design appears plausible: process CO$_2$, loop pressure and
conversion per pass lie inside the published ranges, and the energy indicators lie close to them. The
comparison is only an indication, for three reasons. First, the references leave their boundaries and
conversion factors partly open. The net specific energy
is 26.97~GJ/t with electricity counted as delivered and 30.1~GJ/t at an assumed conversion efficiency of
40\%, against a BAT range of 27.6--31.8~GJ/t ($\pm$1.5~GJ), and the comparison cannot establish whether
the design is more or less efficient than a BAT plant. Second, the design imports more electricity than
the IEA's reference plant (2.1 against 0.3~GJ/t), because all its machines are electric, which lowers
its natural-gas figure, and its lower steam export is not a merit, since its furnace, steam and
refrigeration models are simplified. Third, the comparison was not fully blind, because a design agent
opened the BREF during the run. We found no design value that moved toward the reference figures, but
we cannot exclude influence (SI Section~S11).

\begin{table}[!htbp]
  \centering
  \caption{Key performance indicators of the ammonia design against open references, for plausibility only. Per t of ammonia; lower heating values;
  electricity as delivered (3.6~GJ/MWh) unless stated; exported steam at its enthalpy above make-up water. The references leave boundaries and conversion factors partly open, so
  comparability is limited (SI Section~S11). BAT: best available techniques.}
  \label{tab:kpis}
  \footnotesize
  \begin{tabular}{>{\raggedright\arraybackslash}p{0.3\columnwidth}>{\raggedright\arraybackslash}p{0.22\columnwidth}>{\raggedright\arraybackslash}p{0.4\columnwidth}}
    \toprule
    \textbf{Indicator} & \textbf{Design} & \textbf{Reference} \\
    \midrule
    Net energy, GJ/t & 26.97 & 27.6--31.8 $\pm$1.5 (BREF BAT); 27.6 (IEA BAT) \\
    Net energy, electricity at 40\%, GJ/t & 30.14 & 27.6--31.8 (BREF BAT); 28.05 (IEA, if delivered; our inference) \\
    Gross energy, GJ/t & 30.13 & 32.4 (IEA BAT) \\
    Natural gas, GJ/t & 28.01 & 29.3--31.1 (BREF reported, feed plus fuel, not BAT); 32.1 (IEA BAT) \\
    Process CO$_2$, t/t & 1.164 & 1.15--1.40 (BREF) \\
    Direct CO$_2$, t/t & 1.558 (1.84 with purchased power at 475~g/kWh) & 1.8 (IEA BAT) \\
    Net steam export, GJ/t & 3.16 & 4.8 (IEA BAT) \\
    Loop pressure, bar(a) & 154 & 100--250 (BREF) \\
    Conversion per pass, \% & 25.5 on all gas (30.2 on H$_2$+N$_2$) & 20--30 (BREF, basis undefined) \\
    NOx in flue gas, mg/Nm$^3$ & not modeled & 90--230 (BREF BAT) \\
    NOx factor, kg/t & not modeled & 0.29--0.32 (BREF benchmark) \\
    \bottomrule
  \end{tabular}
\end{table}

\subsection{What the Checks Find and Miss}
\label{sec:evaluation}
Three experiments test the checks and the regeneration of derived files on fresh copies of the reported
state (\texttt{cheme}~1.1.0, DWSIM~9.0.5), and a fourth asks an agent to change the plant. Unmodified,
the state gives 0 errors and 268 warnings and passes 23 of 25 gates. Table~\ref{tab:checks} summarizes
the results; SI Sections~S12 and~S13 list every fault and change.

\medskip\noindent\textbf{Planted faults.}
To see what the checks find, we injected one fault at a time into a copy of the files and ran the
checks. We first wrote 19 faults knowing the rules, aimed at schema, consistency, physical, out-of-date
and review rules, such as a line enlarged
in the line list but not on the P\&ID, or a feed temperature changed without rerunning the simulation.
All 19 triggered their intended rule, with a finding at the edited file in 12 cases and at the edited
line in 10. This confirms that the rules fire on the faults they target, but it says nothing about
faults nobody wrote a rule for. We therefore had a separate agent session write 12 further faults on a
copy of the engineering files from which the checks, skills and reviews had been removed, so that it
could not see the rules. We count a fault as identified when a new finding names the wrong condition
itself; a finding that only says that a file changed or that a result is out of date does not count. By
this definition, the checks identified 4 of the 12: a mistyped relief-valve tag, a vessel placed on top
of another, a hand-edited HAZOP risk band and a trip sharing its transmitter with the control loop. Five
raised only a generic change signal, among them a relief valve set above its vessel's design pressure
and plain carbon steel chosen for a $-33\,^{\circ}$C drum. Three raised nothing, including a let-down
valve changed to fail open. Four further planted values that were implausible but self-consistent, such
as a datasheet pressure off by a factor of ten, also raised only change signals. This means that the
checks find what contradicts an implemented rule, but not an error that no rule covers, even when
another file states the conflicting value.

\medskip\noindent\textbf{Engineering changes.}
To see how far the checks reach, we made five engineering changes on separate copies, from a re-rated
pump to a 20\% capacity increase, and ran the checks once after each. Beforehand, a separate agent
session had listed, for each change, the files it expected to be affected and whether each needed an
edit, a regeneration or a review; no engineer has checked these lists, so the numbers are indicative.
The checks flagged or named 13 of 22 expected files for a re-tagged pump and 5 of 161 for the capacity
increase, and fewer than half for the other three changes. After the capacity increase, the checks
raised balance errors, marked the simulation and 105 stream values as out of date, lowered the
calculation maturity and refused the PFD rebuild, but they did not flag most downstream disciplines,
because the checks stop at the first broken layer until it is repaired. This means that one pass of the
checks shows the engineer where to start, not everything a change will touch.

\medskip\noindent\textbf{Reproducibility.}
We regenerated derived files with fixed software versions. The six generators tested (BFD and PFD,
plant model, one P\&ID rendering, two datasheet drawings, LCA, MTO), two DWSIM runs and library builds
from one commit gave output identical to the byte~\cite{Lamb2022ReproducibleBuilds}. After a
feed-temperature change, a 72~s simulation rerun cleared all 106 out-of-date findings. On another
operating system, the BFD, PFD, plant model and the two datasheet drawings regenerated identically,
whereas the LCA and MTO files differed in their line endings, a defect in the commands that write them.
The derived files are therefore reproducible with fixed software versions, apart from these line
endings.

\medskip\noindent\textbf{Agent-driven change.}
Finally, one unattended agent session was asked to raise the capacity to 1,800~t/d and propagate the
change as far as it could on a budget described as tight, resolving what the checks report and listing
the equipment it judged must be resized, without relaxing a gate or signing off (SI Section~S13). In
49~minutes and for USD~8.55 of a USD~19.90 cap, it raised the design capacity and the feed flows,
re-simulated the plant to 1,799~t/d with closed balances, and updated control setpoints, datasheet
duties, relief loads, the estimate and the LCA; it changed 128 of the 161 files on the expected-impact
list. A separate rerun of the checks found 0 errors and 252 warnings, a calculation maturity of
\emph{balanced}, and four failing gates: the two detail gates that had failed before and two reviews
that the change had made out of date. Yet the plant was not resized. The flows and duties are 20\%
larger on paper, but no P\&ID or line changed, the equipment dimensions and the layout still describe
the 1,500~t/d plant, and 36 lines exceed their design velocities by the agent's own screening, not by
any check, because no rule compares a flow with a line size. The agent reported this and stated that it
would not trust its costs. The tight budget may have contributed, so this result does not measure the
most that agents can do. The outcome is mixed: an agent carried out substantial cross-disciplinary work
through the shared repository, but the checks follow only the dependencies the platform records, so
zero errors did not mean that the change was complete, and a reviewer was still needed.

\begin{table*}[!htbp]
  \centering
  \caption{Summary of the experiments on the ammonia plant (process-library \texttt{nh3-run-final}, \texttt{cheme}~1.1.0); every fault and change is listed in SI Sections~S12 and~S13. Fault counts are
  coverage on these sets. ``Reached'' counts files of an expected-impact list, itself written by an agent, that a finding flagged or named when the checks were first run after the change. These are internal checks and do not establish physical or industrial validity.}
  \label{tab:checks}
  \footnotesize
  \begin{tabular}{>{\raggedright\arraybackslash}p{0.22\textwidth}>{\raggedright\arraybackslash}p{0.11\textwidth}>{\raggedright\arraybackslash}p{0.61\textwidth}}
    \toprule
    \textbf{Experiment} & \textbf{Result} & \textbf{Description} \\
    \midrule
    \multicolumn{3}{l}{\textit{Planted faults}} \\
    Faults written knowing the rules & 19/19 & intended rule fired; these rules work as designed \\
    Faults written without seeing the rules & 4/12 & 5 more raised only a change signal, 3 raised nothing \\
    Implausible but self-consistent values & 0/4 identified & e.g.\ a datasheet pressure off by a factor of ten; only change signals \\
    \midrule
    \multicolumn{3}{l}{\textit{Engineering changes (files reached when first checked)}} \\
    Pump re-rated; re-tagged & 4/19; 13/22 & re-tagging reaches the most; spare, cost and electrical not reached on a re-rating \\
    Line enlarged & 6/14 & HAZOP input and 3D model flagged; list and P\&IDs named \\
    Feed gas changed; +20\% capacity & 4/115; 5/161 & balance errors end the first pass before most later disciplines are checked \\
    \midrule
    \multicolumn{3}{l}{\textit{Reproducibility}} \\
    Same commit, same environment & identical & six generators tested, library builds, simulation \\
    Other operating system (subset) & identical & BFD, PFD, model, two datasheet drawings; LCA and MTO differ in line endings \\
    \midrule
    \multicolumn{3}{l}{\textit{Agent-driven change}} \\
    +20\% capacity, one session & 128/161 changed & 0 errors, but no P\&ID or line resized and 36 lines above design velocity \\
    \bottomrule
  \end{tabular}
\end{table*}

\section{Discussion and Future Perspectives}
\label{sec:discussion}
We return to the problem of Section~\ref{sec:intro}: agents produce engineering work faster than people
can check it. The case shows one way to organize that work and where it still falls short.

\medskip\noindent\textbf{AI acceleration and the practice of engineering.}
In the proposed workflow, agents act on findings, people review a diff with its evidence, and progress
is judged by the evidence instead of the number of drawings. We argue that this is the precondition for
acceleration: without it, faster drafting only produces more unchecked documents. How much of the
acceleration survives review remains unclear. The length of
software tasks that agents complete on their own has doubled about every seven months since
2019~\cite{Kwa2025LongTasks}, and most technology professionals report that AI has increased their
productivity~\cite{DORA2025}, but such gains depend on the task, and this study does not establish a net
time saving once engineering review is counted. In the ammonia case, a package that passed the
basic-stage gates cost USD~224 in API calls for its design sessions, but its reviewers sent work back
13 times, the last gates
needed decisions from a person, and an independent engineer would still have to review the result. We
expect the cost of first drafts to fall much faster than the cost of checking and review, which would
then set the pace and change what teams must build and education must
teach~\cite{Schweidtmann2026PlantAsCode}. Controlled studies against engineering baselines, comparable
to SWE-bench for software agents~\cite{Jimenez2024SWEBench}, are in preparation for process design,
HAZOP and other tasks.

\medskip\noindent\textbf{Known and novel processes.}
Ammonia from natural gas is among the best-documented processes in chemical engineering, and the route
was set in the first task. The case therefore tests whether agents can carry a known process through
engineering work consistently, not whether they can conceive a new one. We believe that particular care
is needed with novel processes, for three reasons. LLMs draw on published knowledge and may fall back on
familiar flowsheets where a different one would be better. New processes lack the kinetic, property and
cost data on which the deterministic tools depend, so more of the design rests on assumptions. And the
rules and gates encode established practice: they can show that an unconventional design is
inconsistent, not that it is good, and they may flag a sound new idea as an error. Cheap and traceable
iterations may still help creative work by letting many more alternatives be generated and screened,
with people judging which are worth pursuing; testing this needs design tasks on processes published
after the model's knowledge cutoff, or hypothetical routes judged by experts.

\medskip\noindent\textbf{Consistent is not correct.}
Keeping calculations outside the LLM makes them repeatable and testable, but a deterministic
calculation with a wrong assumption, an unsuitable property method or an incomplete model gives the
same wrong result every time. The ammonia case illustrated this in part: the energy indicators looked
favorable until their boundaries were examined, a wrong design value that no rule covers raised no
finding that named it, and a capacity change raised no errors although the plant had not been resized.
The agents also wrote some of the calculation scripts themselves. This means that the checks establish
consistency within the dependencies the platform records, and an engineer must still judge the
assumptions, the models and what the rules do not cover. The missed faults point to rules that compare
values that today only change together, such as a datasheet duty against the flowsheet; a rule is a
small piece of deterministic code, and benchmarks with planted faults on open plants would show how far
the coverage has come.

\medskip\noindent\textbf{An open ecosystem.}
Much engineering knowledge sits in proprietary project files, vendor-specific tools and internal
databases. ChemE-as-Code keeps it in open, documented formats, so a project remains readable without
any single application, tools and skills from different organizations can work on the same files, and
commercial systems, whose scope and certification open tools do not replace, can read and write them
through DEXPI and CAPE-OPEN~\cite{Cameron2024DEXPI,BelaudPons2002}. Open projects matter most.
Complete, machine-readable plants with their models, drawings, decisions, safety studies and
inventories have rarely been openly
available~\cite{SchulzeBalhorn2023DataAugmentation,Schweidtmann2024MiningProcessInformation}. They
could serve as benchmarks, teaching material and shared expertise~\cite{Schweidtmann2026PlantAsCode},
and because their assumptions and open points are public, engineers can review them by raising issues
and proposing changes. Life cycle inventories derived from such projects can be recalculated from their
models, which established databases such as ecoinvent~\cite{Wernet2016Ecoinvent} do not expose. The
Process Atlas, an open database of process models and LCIs generated from the process library, is a
forthcoming application~\cite{Schweidtmann2026ProcessAtlas}; its breadth and data quality are not
results of this paper.

\medskip\noindent\textbf{Risks.}
Three risks follow directly from this work~\cite{Schweidtmann2026PlantAsCode}. The first is over-trust.
In the ammonia case, a capacity change produced zero errors although the plant had not been resized; an
engineer who reads a clean check as a complete design is misled. The platform reports calculation
maturity and open warnings next to the errors and records who reviewed what, but the habit of
reading the evidence rather than the status has to be built. The second is liability and certification.
When an agent drafts a design and an agent accepts it, no person has taken responsibility for it, and
current codes assume that one has. Rules are needed on who approves an agent-made design, which record
of the agent's work an approval must cite, and which checks a certifying body accepts as evidence; the
run records, review records and sign-off rules of this platform are a starting point, not a substitute.
The third is the training of engineers. Junior engineers learn much of their craft through the routine
checking and data transfer that agents now take over, and the profession could lose the judgment that
review needs unless this experience is built in new ways, for example by having junior engineers review
and challenge agent work together with its evidence. An error in a shared skill, rule or template could
also spread to many designs at once, and machine-readable plant models raise questions of cyber
security and intellectual property, which we note without treating them here.

\medskip\noindent\textbf{Limitations.}
The case study is illustrative and agent-generated, with the author's recorded interventions. Its
reviewers are agents of the same model as its designers, so their findings show what the model catches
in its own work, not what an engineer would catch. The
checks establish consistency within the rules that exist; engineering judgment and plausibility,
mechanical design, relief sizing, dynamics, the correctness of a HAZOP rating and dependencies not yet
fingerprinted, such as datasheet duties, lie outside them, which is why the held-out faults and the
capacity change found what they did. The case uses one general-purpose agent in workflow roles and
offers no controlled comparison with other agent architectures or with the same agent without skills,
the comparison that would show what the skills contribute. Proprietary tools hold most industrial
data, DEXPI coverage is incomplete, and agents are nondeterministic, so two runs of the same brief give
different designs. ChemE-as-Code is a first open research prototype, and these limits should be read in
that context. The same pattern could extend
to the operation of existing plants, to laboratory work where experiments and synthesis procedures are
already written as code~\cite{Aguilar2024ExperimentsAsCode,Mehr2020ChemicalSynthesis}, and to molecular
and product design; the core is layered by domain so that such domains can be added.

\section{Conclusions}
\label{sec:conclusions}
AI agents offer the prospect of substantially accelerating chemical process engineering and reducing its
cost, but realizing this requires more than capable models. We argue that engineering knowledge must be
connected, machine-readable and checkable, so that agents can work across disciplines while engineers
keep responsibility for decisions and approval.

ChemE-as-Code implements this through two foundations: a versioned repository of linked engineering
files, and an open library of deterministic tools and agent skills that spans chemical engineering from
catalysis and reaction kinetics to plant layout and LCA. A general-purpose agent uses these across disciplines, while
calculations, checks and approval stay outside the LLM, with approval reserved for people in the
proposed workflow. In an illustrative natural-gas ammonia case, agent sessions turned a written brief
into a preliminary package that passed the platform's basic-stage gates, with HAZOP, datasheets and
relief cases, for USD~224 of API cost in the design sessions, and a comparison with open references indicates that the
design is plausible. The checks caught every fault written with knowledge of the rules but only some of
the faults written without them, and a capacity change raised no errors although the plant had not been
resized. An error-free result therefore shows consistency, not a complete or correct design. The case
shows that the workflow runs. It does not establish the quality of the design or a productivity gain,
and no independent engineer has reviewed the result.

Beyond automation, ChemE-as-Code lays a foundation for an open chemical engineering ecosystem in which
process models, engineering data, decisions and the evidence behind them are reusable assets rather
than remaining inside single tools and projects. The longer-term opportunity is to change how
engineering knowledge is created, checked, shared and reused, first in process and plant design and
then across chemical engineering.
\section*{Data and Code Availability}
ChemE-as-Code is developed openly in the ChemE-as-Code organization on GitHub
(\url{https://github.com/ChemE-as-Code}); all repositories listed below are public.
The core library, which validates, formats, renders and compares DEXPI P\&IDs and engineering
YAML files kept in Git and includes the web viewer for plant repositories, is available at \url{https://github.com/ChemE-as-Code/core} (GNU AGPL-3.0-or-later).
It builds on pyDEXPI, available at \url{https://github.com/process-intelligence-research/pyDEXPI}.
A template for starting new plant
repositories is available at \url{https://github.com/ChemE-as-Code/project-template}; the template carries no license,
so that the content of each plant repository remains its owner's to license.
The process library, including the ammonia case study (\texttt{processes/ammonia-1}) and its run
records at tag \texttt{nh3-run-final},
is available at \url{https://github.com/ChemE-as-Code/process-library} (code under GNU AGPL-3.0-or-later,
except the LCI database installer, which is under the MIT license; process data and designs under
CC~BY~4.0 unless a process states otherwise).
The project website is \url{https://cheme-as-code.org}, and the Process Atlas generated from the
process library is available at \url{https://atlas.cheme-as-code.org}. The scripts and raw results of the evaluation in Section~\ref{sec:evaluation} are archived on Zenodo~\cite{Schweidtmann2026Zenodo}.

\section*{Declaration of AI Use}
The author used GPT-6 (OpenAI) and Claude Opus 5.5 and Claude Fable 5.1 (Anthropic) to assist with drafting, language editing, literature searches and figures. The author reviewed and revised all content, verified the sources and takes full responsibility for the manuscript.

\section*{Acknowledgments}
This research is supported by the Netherlands Organisation for Scientific Research (NWO) under the
Veni grant (ID 20196), for which we express sincere gratitude. The author thanks the current and former members of the Process Intelligence Research Group, whose work over the past five years contributed to this paper. In particular, the author thanks Sophia Rupprecht and Qinghe Gao for their prior contributions on multi-agent systems for separation process design.

\section*{Conflict of Interest}
The author declares no potential conflict of interests.

\bibliographystyle{unsrtnat}
\bibliography{references}

\clearpage
\includepdf[pages=-]{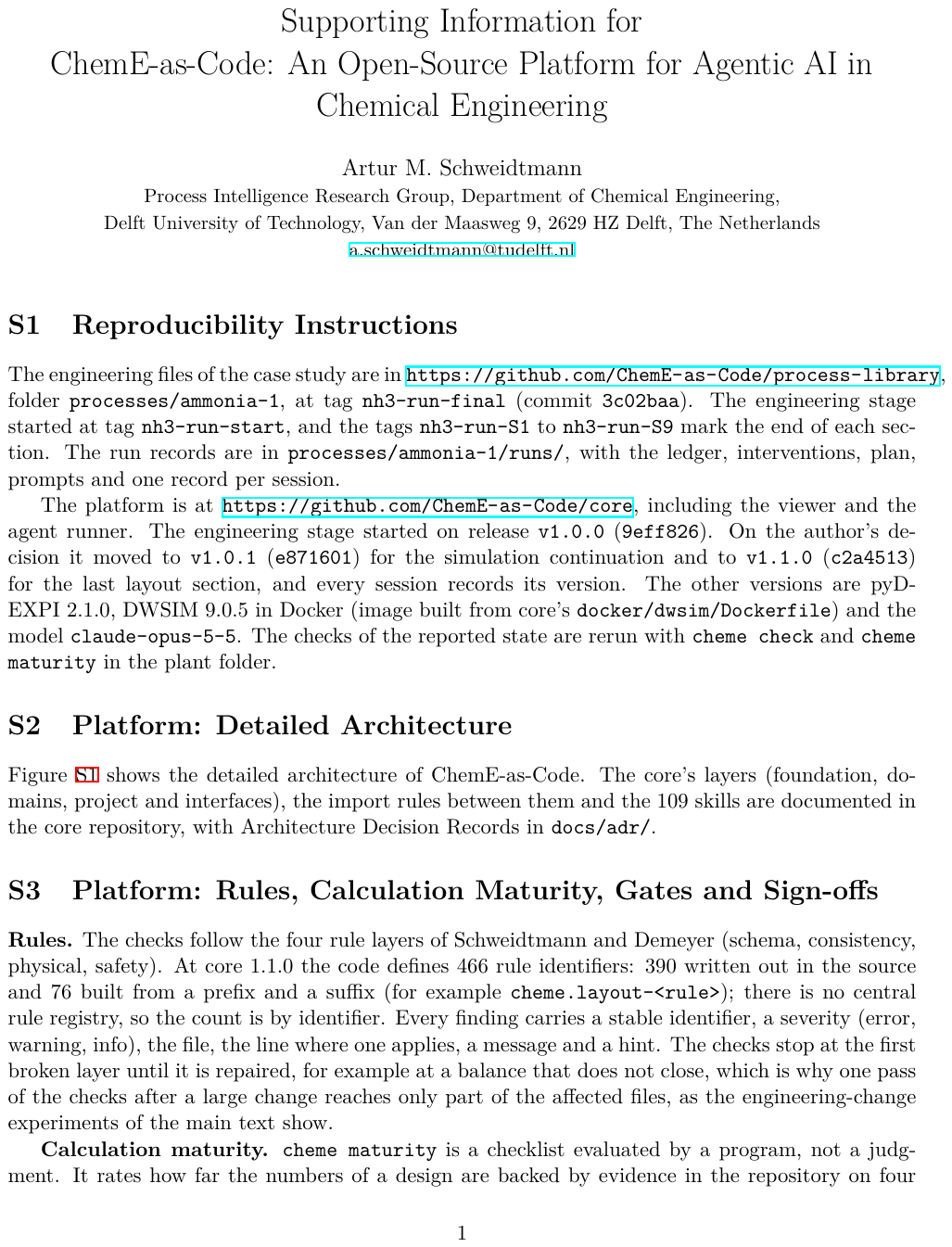}
\end{document}